# Brain-language fusion enables interactive neural readout and in-silico experimentation

**Authors:** Victoria Bosch[1*], Daniel Anthes[1], Adrien Doerig[1,2], Sushrut Thorat[1], Peter König[1], Tim C. Kietzmann[1*]

**Affiliations:**

[1]Institute of Cognitive Science, University of Osnabrück, Osnabrück, Germany.
[2]Department of Psychology and Education, Freie Universität Berlin, Berlin, Germany.

*Corresponding authors: vbosch@uni-osnabrueck.de, tim.kietzmann@uni-osnabrueck.de

## Abstract

Large language models (LLMs) have revolutionized human-machine interaction, and have been extended by embedding diverse modalities such as images into a shared language space. Yet, neural decoding has remained constrained by static, non-interactive methods. We introduce CorText, a framework that integrates neural activity directly into the latent space of an LLM, enabling open-ended, natural language interaction with brain data. Trained on fMRI data recorded during viewing of natural scenes, CorText generates accurate image captions and can answer more detailed questions better than controls, while having access to neural data only. We showcase that CorText achieves zero-shot generalization beyond semantic categories seen during training. In-silico microstimulation experiments, which enable counterfactual prompts on brain activity, reveal a consistent, and graded mapping between brain-state and language output. These advances mark a shift from passive decoding toward generative, flexible interfaces between brain activity and language.

# Introduction

The emergence of large language models (LLMs) has transformed human-machine interaction by harnessing natural language as a flexible medium for reasoning and integrating knowledge across diverse domains. Extending this paradigm beyond text, recent advances in visual and multimodal models fuse image and video data into language embedding spaces, allowing open-ended textual access to complex sensory information[1,2]. By translating diverse sensory inputs into a shared language space, these models enable cross-modal reasoning and flexible generalization, turning sensory data into a format accessible through natural language.

By contrast, machine learning applications in neuroscience, in particular for the decoding of neural signals, are constrained by remaining non-flexible, and non-interactive. This is most obvious for cases in which experimental conditions or categories are read out from neural data[3–6], but the same limitation holds for more recent approaches that map neural data into LLM embedding spaces, for example to decode image captions[7–11]. Irrespective of the target format (categorical or continuous), stimulus format (dynamic or static), and data format (time-resolved, or temporally averaged), prior decoding approaches rely on the ability to train a mapping between brain responses and static, predefined output vectors. As a result, different questions about the neural code usually need different decoders, each requiring sufficient data and labels for the target dimension to be examined, thus hindering generalisation beyond the training data. In contrast, learning a mapping directly into a generative natural language model, may overcome these limitations by providing open-ended linguistic access to brain scans and enabling continuous probing of the brain-language mapping. A successful interface with LLMs thus holds promise for flexible readout of information from one decoder, generalisation to new domains given little data (zero-shot decoding of language-accessible concepts, or in-context learning), the possibility to integrate data from different datasets and imaging modalities into one common framework, and direct access to the full capabilities of LLM agents, including reasoning, and integration of background knowledge.

Here, we demonstrate a first step towards creating such an open-ended interface with brain data. To move beyond prior representational analyses, we propose a novel decoding framework, CorText, that fuses neural data directly into the latent space of a large language model[12] to create an interactive, natural language interface to brain activity that enables both, more traditional decoding analyses, as well as novel modes of open-ended language-based interaction with the data. As a proof of concept, we apply CorText to a large-scale 7T fMRI dataset collected while participants viewed natural scenes[13]. We demonstrate that the system is capable of generating accurate captions of the images shown to the participants while solely relying on the corresponding brain data, without ever having access to the actual stimuli. Furthermore, CorText successfully answers flexible, open-ended questions, which the system answers by jointly engaging with the brain embeddings and language. Generalisation to concepts not encountered during training is shown by zero-shot generalisation experiments. Finally, we introduce a new counterfactual analysis method that emulates the effects of cortical microstimulation in silico, while at the same time enabling language-based probing of the resulting information. Together, these advances mark a shift from non-interactive, rigid neural decoding toward a unified, generative, and flexible framework for interfacing with brain data.

# Results

## Interactive, language-based neural readout from fMRI data

To enable flexible neural decoding and diagnostic probing of visually evoked neural responses to complex scenes, we developed CorText, an end-to-end trainable multimodal model, bridging fMRI data and language. Central to the approach is the idea to learn a nonlinear mapping from parcellated brain data into the embedding space of large language models. As a result, each of the activity patterns observed in a given brain region are translated into variable tokens, which are prepended to the language prompt. This encoding approach thus translates brain data into language-compatible tokens, enabling the language model to access the information encoded in the brain and subsequently to integrate it into the language interface for conditional language generation (Fig. 1A).

To fuse brain data into the language embedding space, the whole brain is first parcellated into functional components[14] (Schaefer parcellation; see Fig. S1 for exploration of different parcellations). All vertices of each parcel are then run through a region-specific brain tokenizer, i.e. a neural network that learns to map the distributed brain responses into language embedding space (see Methods). Using region-wise brain tokenization preserves spatial (functional) structure and allows for use of the language decoder's attention mechanism. Once translated into language space, the brain embeddings can be concatenated with the input question embeddings, and the resulting multimodal input token sequences are fed into a pretrained large language model, here Llama 3.1[12], which autoregressively generates answers conditioned on the neural context (Fig. 1A). Following prior multimodal alignment approaches[1,15,16], training consists of two phases. In the first phase, only the brain tokenizers are trained, while the language model remains frozen. In the second phase, we apply QLoRA, a parameter-efficient finetuning method, to additionally train a low rank adapter on the attention matrices of the language decoder, while keeping decoder weights frozen and thus retaining the language decoders abilities[17].

We apply CorText to the Natural Scenes Dataset (NSD), a large-scale 7T fMRI dataset of neural responses to images of complex natural scenes[13]. Using the data of four separate participants who participated in the whole experiment, CorText is trained to predict image captions and instruction question-answering pairs available for the stimuli shown to the participants[18] (see Methods). Importantly, at no point in time does CorText have direct access to the images shown to the participants, but is solely provided with the brain data collected while the participants observed the stimuli.

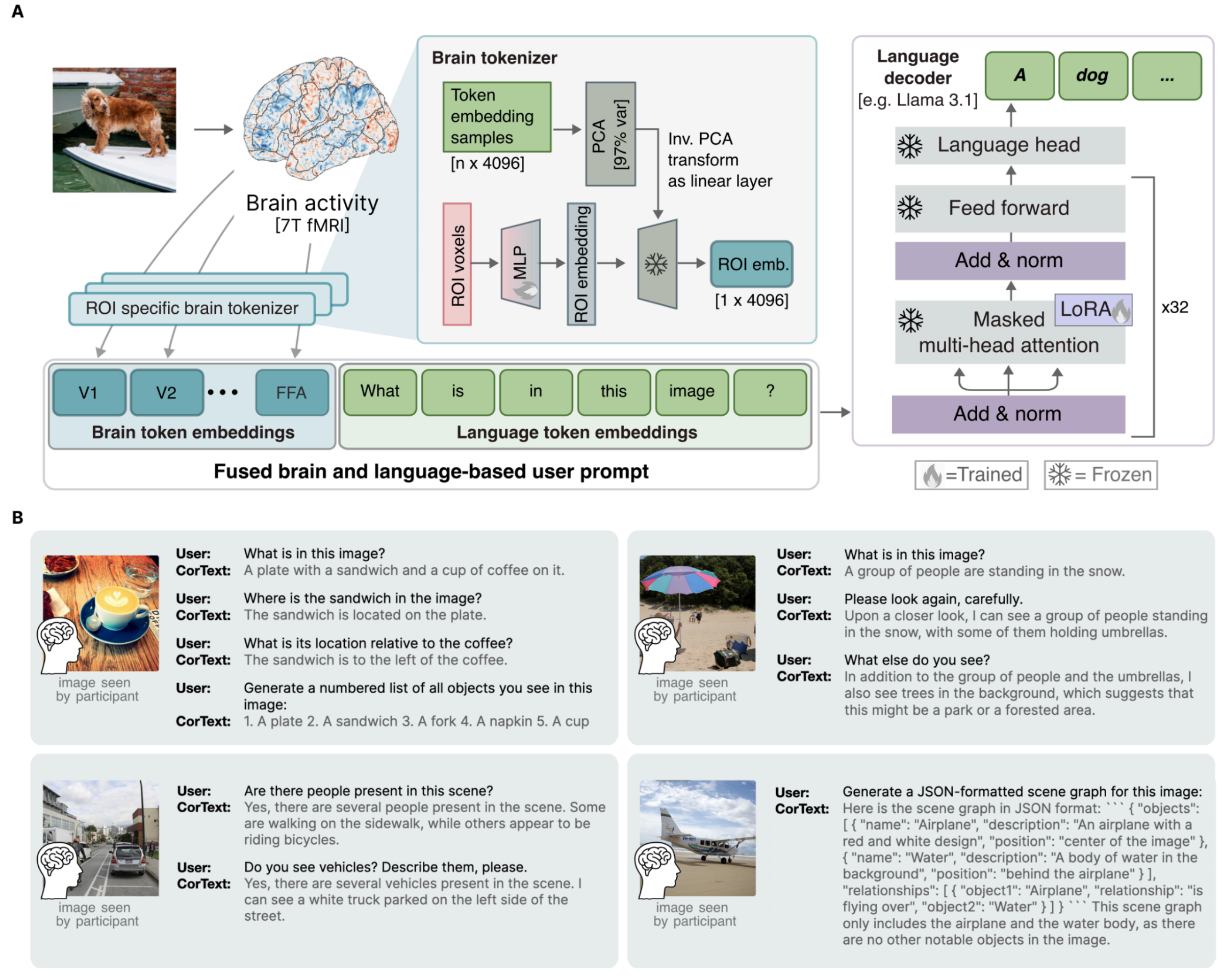

**Fig. 1. CorText enables interactive neural decoding.** (**A**) Visualization of CorText pipeline and architecture. Region-specific neural responses to natural scenes are translated into the large language model's embedding space by end-to-end trainable brain tokenizers. The resulting tokens are fused with text prompts, and passed into a pretrained language decoder to generate natural language outputs. **(B)** Example conversations with CorText, here shown for participant 1, on unseen test trials reveal its versatile neural decoding capacities. Because CorText embeds neural data directly into the latent space of a pretrained generative language model, it inherits its broad linguistic capabilities, supporting, for example, multi-turn dialogue (allowing follow-up questions that probe additional aspects of the perceived scene, or requests to reason more carefully yielding revised answers), structured outputs (e.g. JSON-formatted scene graphs), or prompts for object lists.

We trained one CorText model instance for each participant, and tested performance on held-out neural test trials consisting of unique scene IDs shown to all participants (~5% of available neural data. Conversations with CorText illustrate its capacity to read out information from the brain data in a highly versatile manner (see Fig. 1B for examples of participant 1), demonstrating successful generalization to several forms of task and language outputs not seen during training. To quantify the model's general ability to extract semantic stimulus information from the brain data, we first evaluate its capacity to provide language descriptions of the perceived scene images. We prompted the system with, for example, "*Give a concise and descriptive caption of this image:*", or "*What's in this picture?*", and evaluated its answers against human scene captions for the corresponding natural scenes (CLIPScore, see Methods). Despite only having access to brain data, and not to the stimulus images themselves, CorText reliably generated scene captions that closely matched the human captions for the same images (Fig. 2A; see Methods and Supplemental tables S1-5 for further examples; results shown for participants that have completed the NSD experiment, evaluation for all subjects can be found in Fig. S2). Interestingly, if errors occurred, the system often mistook semantically or visually similar semantic categories for one another (e.g. surfers were confused with skateboarders or skiing/snowboarding). The mode of the performance curves (0.81 for subject 1) reached 92.75% of the mean human consistency (0.87). As a control model, we trained CorText on entirely shuffled brain data, effectively removing all neural information and testing the model's performance when only relying on prompt information and the language model's prior knowledge. CorText significantly outperformed this control (performing at 0.46; Welch's one-sided t-test, all $p<0.01$), which tended to output roughly the same single sentence across test trials. These results demonstrate CorText's ability to learn a mapping between the high-dimensional neural activity and the language embedding space of a large language model, thereby enabling the interface to natural language and the generation of language outputs derived from brain signals.

Next, we assessed performance on more varied instructions. To do so, we evaluated CorText's answers to fine-grained visual questions from the LVIS-Instruct4V dataset[19], again on test trials held out during training. This dataset includes, among others, questions about numerosity, object positions, select subparts of the scene, or object presence. The system's output and ground-truth answers were compared by calculating sentence-embedding cosine similarity using QWEN-3 8B[20] (see Methods). Across participants, CorText consistently and significantly outperformed baseline alignment, estimated via shuffling trial ids, as well as the trained control model (Welch's one-sided t-test, all $p<0.01$), for example by producing answers that better captured the semantic content of the perceived images (Fig. 2B), or more accurate estimates of numerosity (see Fig. S3). Compared to the semantic captioning task considered first, which considers the scene as a whole, the request for more detailed answers revealed the system's ability to dynamically focus on different aspects of the scene given the prompt (see examples in Fig 2B, right; see Supplemental tables S6-9 for more examples).

Together, these examples highlight that fusing brain data into a generative language model yields not only accurate decoding, but a flexible, task-general interface for interrogating neural representations through natural language.

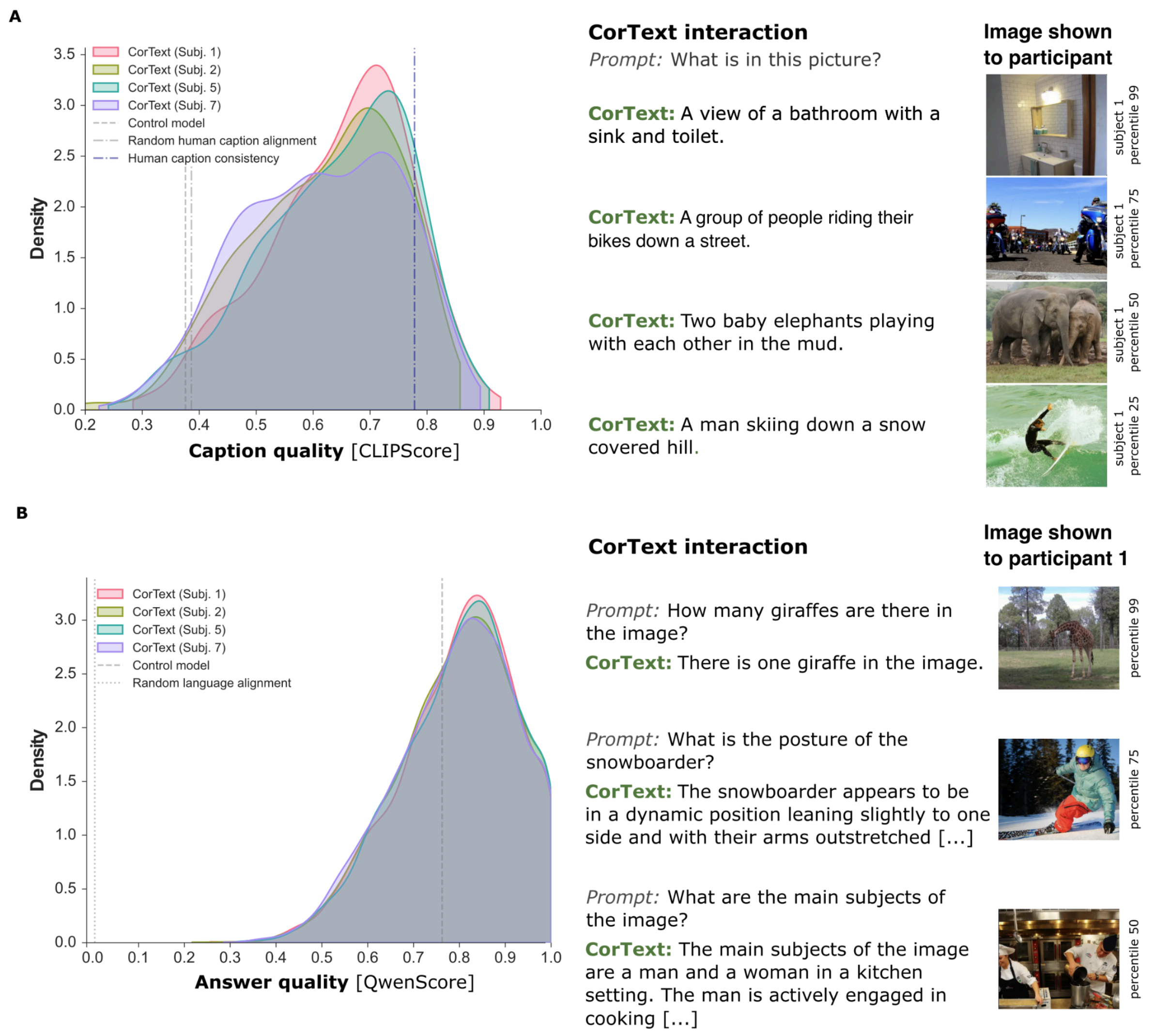

**Fig. 2. Quantitative evaluation of CorText performance.** (**A**) Distribution of caption quality (CLIPScore) for captions generated from fMRI data, compared to two control conditions: a model trained based on shuffled data, as well as human caption consistency. Example outputs of CorText trained on data for subject 1 illustrate variability in description quality at different percentiles of scores. (**B**) Performance on fine-grained question-answering (QwenScore) by CorText, together with two control conditions: a model trained based on shuffled data, as well as baseline alignment estimated via shuffling trial ids. Example outputs show CorText's ability to capture semantic detail beyond broad captions.

## Zero-shot neural decoding

To test if CorText can suitably generalize to novel concepts not encountered during training, we trained additional CorText model instances for which we held out all trials corresponding to a given semantic concept. As reported by Shirakawa et al.[21], the train and test images of NSD share semantic content (although images remain unique between splits), such that CLIP feature translators, often used in neural image reconstruction, are incapable of generalizing to novel semantic clusters. This observation underlines the importance of this analysis, as it assesses the degree to which CorText may be overfitting on the semantic clusters of the training set. The analysis furthermore highlights the system's ability to meaningfully answer questions about concepts that the system was not explicitly trained for.

Three CorText models were trained on data of participant 1, using three modified datasets in which all trials depicting zebras, surfers, or airplanes were excluded, respectively. These categories were chosen after inspection of the UMAP embeddings of the dataset captions (Fig. 3A,C), indicating that these categories are located in the data periphery, rendering the task more challenging. After training, each model was tested on the held-out trials by asking the model to provide image captions to the previously unseen concept (Fig. 3B). Explicit mention of the zero-shot category names was not expected as the corresponding tokens were withheld from training. However, the predicted zero-shot captions clustered in close vicinity of the corresponding ground-truth captions for the held-out categories, demonstrating that CorText can meaningfully generalize to semantic categories never seen during training, while also correctly describing the spatial layout of the scene (Fig. 3C; example captions for each model in Fig. 3D; see tables S10-12 for more examples). For example, when showing an image of a zebra for the first time, the system would describe it as a horse, or giraffe, indicating that the system correctly classified the shown images on a higher level of abstraction (e.g. four-legged animal), and integrated knowledge about known background settings into its output (e.g. savanna).

As CorText can generalize to novel types of language tasks, we moved into a more explicit question setting (alternative forced choice), and asked the zero-shot trained systems to decide whether a shown brain scan was collected while the participant viewed a zebra, surfer, or airplane. Responses were categorised as indicating one of the three categories, as well as "non-compliance with the task". Despite never having seen the corresponding semantic categories, the system exhibited a high likelihood of choosing the correct target category (Fig. 3E; see tables S13-15 for model outputs). Presented with the held out semantic category, CorText$_{\text{w/o zebra}}$ and CorText$_{\text{w/o surfing}}$ chose the correct response in all task-compliant trials, and CorText$_{\text{w/o airplane}}$ in 58.3% of cases (chance being 33.3% in case of task-compliance). Total rates of non-compliance with the task were low (8.3%, 11.1%, and 5.6% for the three respective models). To ensure that the models are not simply strategizing by picking the category they are least confident about, we also train a CorText model with all three categories held out. This model chooses the correct category with 96,1% (zebra), 69,2% (surfing), and 58,3% (airplane) accuracy for the respective test trials (see Fig. S4).

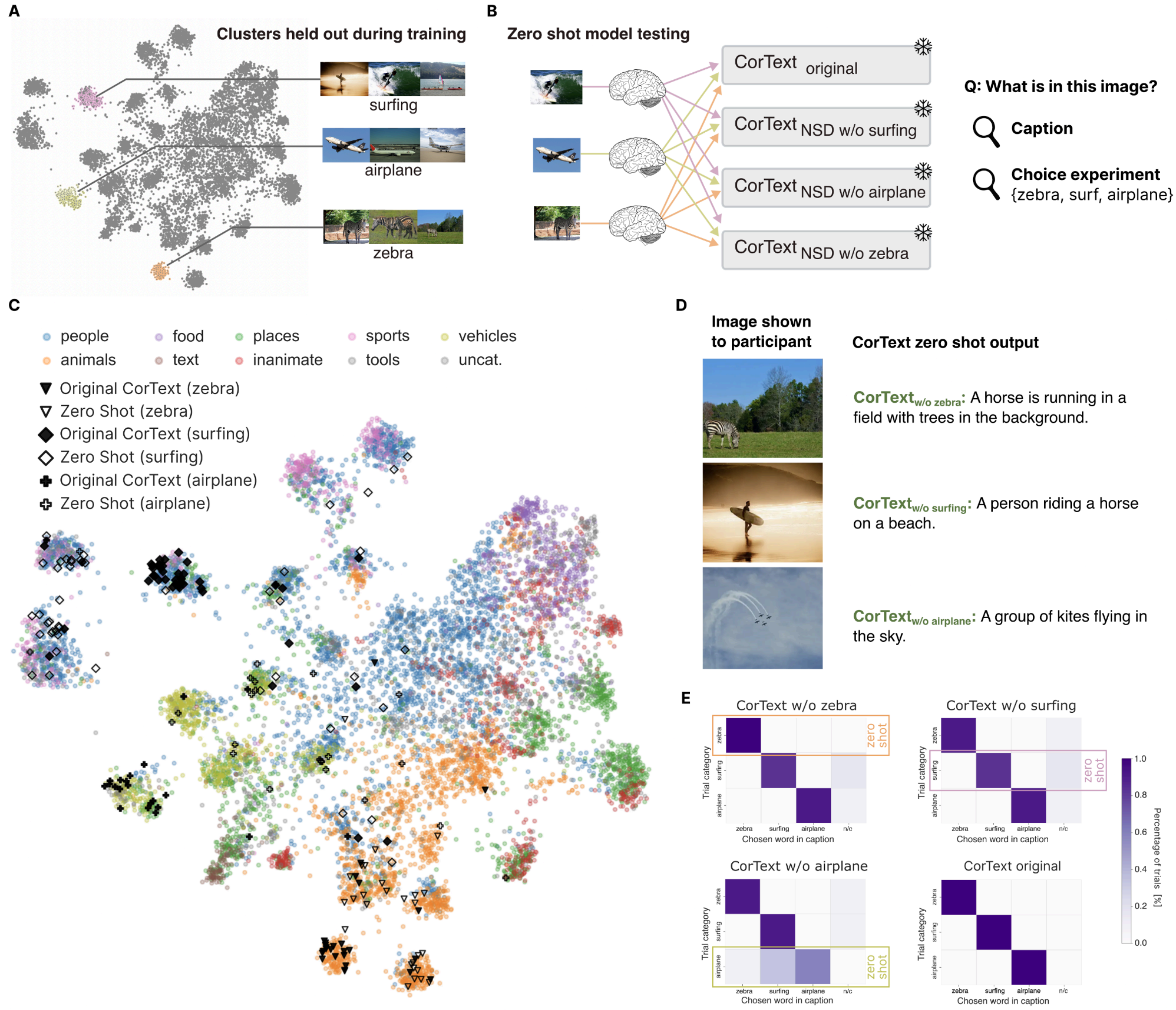

**Fig. 3. Zero-shot neural decoding.** (**A**) Training three CorText models with one entire semantic category cluster held out from the training data. Clusters were chosen to lie at the periphery of the data. (**B**) Evaluation setup for zero-shot generalization on test trials of the held-out category included two test cases: free captioning, and a more targeted choice experiment. (**C**) UMAP projection of CLIP embeddings shows that generated captions for held-out categories cluster near ground-truth human captions (coloured by semantic supercategory), indicating meaningful semantic generalization. (**D**) Example zero-shot caption generations for each model. (**E**) An alternative forced choice experiment confirms the zero-shot models' ability to successfully detect and classify unseen categories.

## In-silico microstimulation

Beyond supporting interactive and flexible decoding, CorText also enables counterfactual analyses, in which researchers can intervene directly on brain data and test, in an open-ended manner, for the resulting effects on the language-based output. This approach is akin to microstimulation (or lesioning) experiments in which researchers test for causal effects of brain activity in a select region via targeted excitation (or inhibition). While the effects of any brain activity perturbation can be studied using this approach, we here use it for an important sanity check: will CorText respond in a semantically coherent manner if we perform in-silico microstimulation of the brain data in terms of known patterns of categorical selectivity, e.g. by stimulating (or inhibiting) face-selective regions[22–24]?

For these experiments, we rely on localizer data from the Natural Scenes Dataset to identify vertices that are preferentially responsive to the presence of faces in the stimulus materials. In-silico microstimulation targets were selected as the top 1% of the most face selective vertices (estimated via the corresponding vertex t-values). This particularly small fraction of data equates to a highly precise, yet distributed microstimulation target. Excitatory in-silico microstimulation was performed by increasing the value of the targeted fMRI betas for a test trial (weighted by a microstimulation strength parameter beta). Inhibitory microstimulation was performed by a corresponding decrease (see Methods, Fig. 4A).

When presented with microstimulated data, CorText indeed demonstrates highly systematic changes in its behaviour. Applied to data of scenes without people, excitatory microstimulation increased the likelihood of person-related terms in both the language decoder outputs (prompted for scene captions), as well as in the model's internal evidence for people-related words as measured by change token probability (Fig. 4B; example captions provided alongside the quantitative data; see table S17 for more examples). Conversely, applying negative microstimulation to scenes that originally included people suppressed such terms, effectively "removing" reference to people from the decoded description (Fig. 4C; see table S18 for more examples). Excitatory as well as inhibitory effects scaled with microstimulation strength, demonstrating graded control over the semantic content expressed in language. We replicate these results by stimulating or inhibiting vertices that show higher activations for car stimuli than other categories, resulting in increased or decreased evidence for vehicles (see Supp. Fig. 6, table S17 and table S18). Applying microstimulation with a randomly selected mask does not show the same effect, confirming that the location of stimulation is indeed causally relevant in inducing or removing visual content from CorText output (Supp. Fig. 7). These results act as an important proof of concept for the potential of brain-language fusion in hypothesis-driven in-silico perturbation experiments on brain representations.

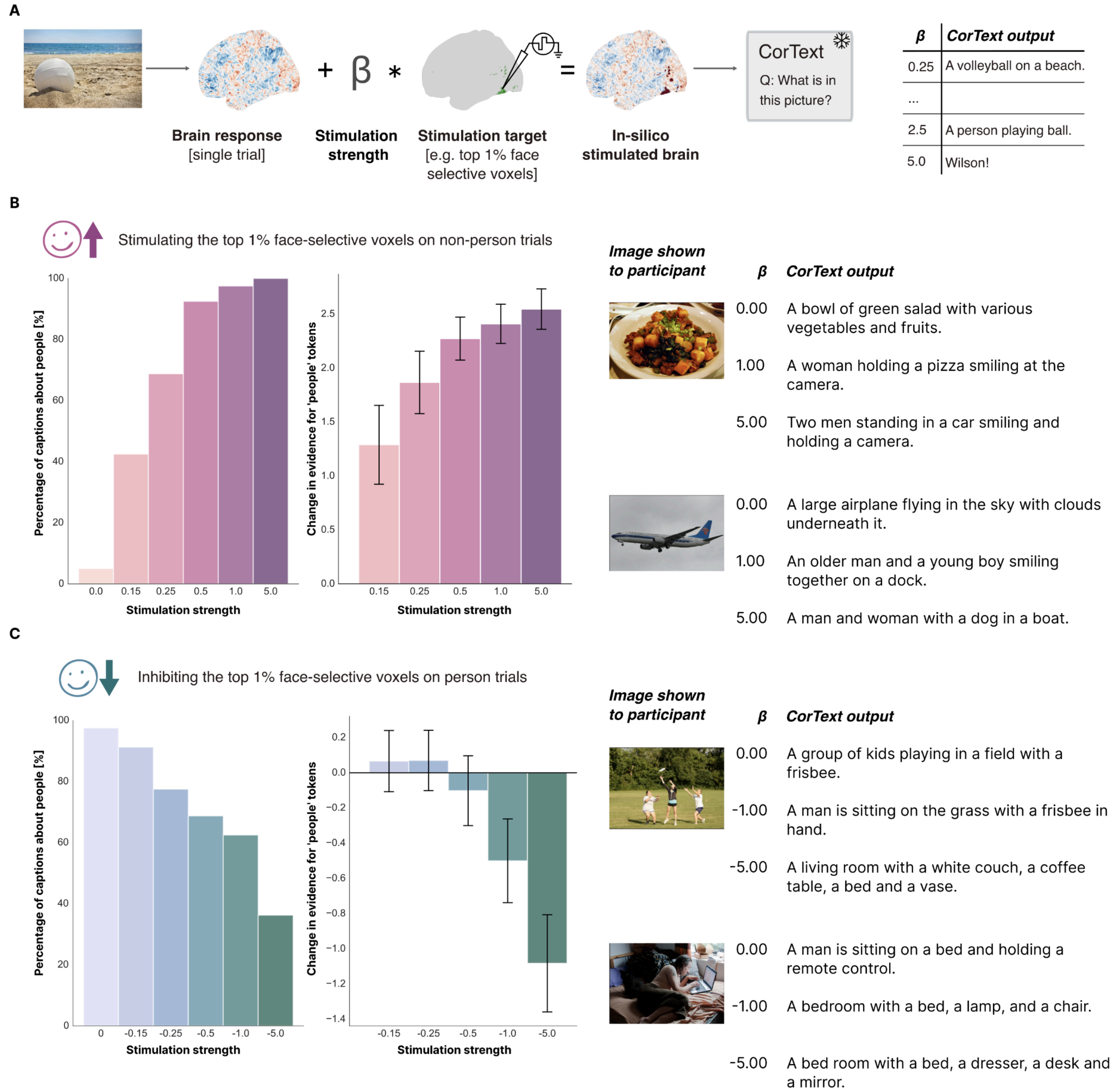

**Fig. 4. In-silico microstimulation.** (**A**) Conceptual illustration of our in-silico microstimulation setup**.** Trial fMRI responses are perturbed by stimulating or inhibiting vertices, here selective for faces. The resulting stimulated response is passed to a trained CorText model together with a captioning prompt. (**B**) Microstimulation of top 1% of face vertices in non-person trials increases the percentage of captions mentioning people (left), as well as the evidence for 'people' tokens (middle). (**C**) Inhibiting the top 1% of vertices with strong responses to the presence of face in trials showing people decreases the percentage of captions mentioning people (left), as well as the evidence for ‘people’ tokens (middle). Example captions (right) illustrate gradual semantic shifts induced by perturbation.

## Discussion

In this work, we introduced CorText, a novel framework for brain-language fusion. By integrating fMRI data into the latent space of large language models, this approach enables open-ended language-based interrogation of brain activity. Focusing on the domain of vision, we demonstrated that the approach enables flexible decoding of varied information about perceived visual content from fMRI data alone, that it can perform zero-shot generalization to novel concepts, and that it enables in-silico experimentation and open-ended language-based probing.

Decoding of neural content has significantly advanced in recent years, with new machine learning techniques, and in particular, language and vision models, fueling its development[7–9,11,25–27]. An important insight from this line of research is that the structure of natural language is not only convenient for human use and interpretability, but also provides a powerful scaffold for uncovering structure in neural data as well as other data modalities[7,15]. This perspective reflects the idea that different modalities may (partially) share latent geometries. Furthermore, natural language has rich compositional structure, and thus likely offers an especially effective scaffold for alignment with high-dimensional (neural) representations, leading to successful integration with different modalities into a unified semantic space through fusion.

CorText builds on these insights. Whereas previous approaches map neural responses onto static target vectors, CorText learns to embed brain activity directly into the latent space of a generative language model, transforming decoding from a passive readout into an interactive interface that can flexibly interrogate the semantic content of neural activity. This shift from passive readouts to generative, interactive engagement with brain data marks a conceptual advance in how we approach the study of neural data and integrate it into AI systems.

Methodologically, CorText differs from previous approaches in important ways. We provide an end-to-end training pipeline that neither includes access to the underlying images during training, nor pre-trained vision-language models that have seen the underlying stimulus materials. This reduces the risk of memorization, which is a central challenge for neural decoding on the Natural Scenes Dataset due to the substantial semantic overlap and visual similarity between training and test sets[21]. Our analyses demonstrate that CorText exhibits the capacity for generalization of higher semantic categories and spatial compositions to categories of trials that were not part of training, revealing a continuous and successful mapping between neural activity and the semantic space of the (brain-fused) LLM. In addition, our approach supports counterfactual perturbations (i.e., asking 'what if'-questions about neural data), opening avenues for interactive in-silico experiments as a new framework for discovery in neuroscience[28]. While the current approach marks an important first step, we expect future work to use advanced mechanistic interpretability techniques to further understand the extent and form of the learnt brain-language alignment.

Future work will explore ways in which brain-language fusion of CorText can further benefit from the broader range of capabilities that LLMs offer, including reasoning, in-context learning, and the integration of web-searches and additional background knowledge. Moreover, it will be of interest to explore how fusion with brain data can enhance capabilities of LLMs[29,30]. Because of the modularity of the CorText architecture, future work will also include systematic testing of different architectural components, such as the language backbone and brain tokenizers. An important direction of research will furthermore explore generalisation to (a) other sensory modalities, (b) to new participants[11,31] (c) the application to other neuroimaging modalities, such as M/EEG, (d) the use of CorText to integrate various datasets, and prior scholarly knowledge, into a joint foundation model of brain-language alignment. If successful, these developments may not only provide powerful ways of probing brain data, but may also interpret and contextualise the underlying activity patterns within broader scopes of neuroscientific knowledge. Together, these exciting directions highlight that CorText lays the foundation for a new, language-based approach in neural decoding.

# Methods

## Data

### Neural data

The Natural Scenes Dataset (NSD) contains 7T fMRI measurements of 8 participants who have each viewed up to 9000 unique images and up to 1000 shared images from the MS COCO dataset[18]. Out of all 8 participants, 4 have not finished the NSD experiment completely, resulting in few available training trials. Hence, we depict the results for the 4 participants that have completed it in the main text (see table S1 for results of models trained on other participants). We use the beta values resulting from the general linear model of the 1.8-mm volume preparation in fsaverage space and average the three stimulus repeats. We split the shared trials into a test and validation set, where the test set contains 515 shared trials. Only trials with available captions and question–answer pairs were retained. (See *Language datasets*). The neural data was parcellated into regions of interest (ROI) using the Schaefer atlas, using the 100 region parcels of the 7-network clustering[14]. We also trained models with higher-resolution Schaefer parcellations (200–500 regions) and the HCP-MMP1 atlas, comprising 360 regions[32] (see Fig. S1).

### Language datasets

Each image in NSD is paired with five human captions from MS COCO. We used these human captions to construct question prompts for scene description. The question prompts are randomly selected from the following:

*"Give a concise and descriptive caption of this image:"*
*"Describe the following image in detail in one sentence:",*
*"Provide a detailed description of the given image in one sentence:",*
*"Describe the important features of the scene in this image:",*
*"What is in this picture?".*

In addition, we included up to 10 question-answer pairs per trial from the LVIS-INSTRUCT4V visual question-answering dataset[19], which contains ~220K question-answer pairs for MS COCO. We extract the pairs that are present for the trials used in NSD.

## Architecture and training

**Language decoder**

CorText is a multimodal decoder-only transformer-based architecture using Llama 3.1 8B-Instruct[12] as the backbone. Llama 3 is an instruct-finetuned generative language model, which has been finetuned using supervision and reinforcement learning from human feedback (RLHF). We use the Hugging Face implementation , and keep the model weights quantized in 4-bit precision for memory efficiency. After embedding the neural data for a trial using these ROI-wise brain tokenizers (see *Brain tokenizers*), the *n* ROI brain embeddings are prepended to the Llama tokenizer-embedded question. The sequence is formatted into Llama's prompt structure with instruction tokens, yielding a multimodal input sequence of brain and text embeddings. Results for a CorText model based on Llama 3.2-1B-Instruct have been included in Fig. S5.

**Brain tokenizers**

To enable multimodal fusion between neural data and language, we use region-wise brain tokenizers, one per ROI defined by the used parcellation. These are implemented as single-hidden-layer MLPs, initialized with Torch's default Kaiming weight initialization[33], All training and experiments are initialized with an identical random Numpy, Huggingface and Torch seed. To reduce parameter count, we use a low-rank linear projection within each tokenizer. Llama-3.1-8B has a 4096-dimensional token embedding space, and as a result, training tokenizers to map into this space would result in a large number of trainable parameters in the last layer of the MLP. Instead, we first apply principal component analysis (PCA) to the 4096-dimensional Llama-3.1-8B word embeddings of all training captions of subject 1 to project onto the first 921 principal components, which capture 95% of the variance. We then append the transpose of this projection as the last linear layer with fixed parameters to each tokenizer such that the remaining layers learn the encoding in the lower-dimensional space. This low rank projection is re-used for all CorText models. The MLPs map the brain data from each region from the number of ROI vertices into the 921-dimensional space by projecting via a hidden layer of input size 460 with a GeLU non-linearity.

**Training**

Model training consists of two phases: pretraining (20 epochs) and finetuning (2 epochs). During pre-training, only the brain tokenizers and layer normalization throughout the Llama language decoder are trainable, effectively training a brain-tokenizer compatible with the language decoder. In the finetuning phase, we employ Quantized Low-Rank Adaptation (QLoRA[17]) to adapt the query and value projection matrices of the decoder, while keeping the base decoder frozen. We set the QLoRa rank to 16, alpha = 16, and dropout = 5e-2. Optimization used AdamW-8bit[34] with batch size 5, learning rate 1e-3 (pretraining) and 2e-5 (finetuning), cosine learning rate scheduling, and L2 regularization on the brain tokenizer weights (0.2 pretraining, 5e-4 finetuning). In both phases, the model is trained to minimize cross-entropy loss between generated and target answers to questions based on the brain data 'context'. We train one CorText model per NSD participant (n=8), and one control model on the training data of subject 1, shuffled both across trials and vertices, which effectively removes all trial-related neural information from the prompt while holding the final prompt length constantand testing the model's performance when only relying on prompt information and the language model's prior knowledge.

### Generation and inference

During inference, answers to questions in the test set are generated using beam search (2 beams), nucleus sampling with a minimum probability threshold of 0.2, and a temperature of 1. For the in-silico experiments, we directly inspected token probabilities and output captions for answers generated using a single beam.

Models are custom made using Python 3.10, PyTorch (version 2.2), making use of Huggingface (version 4.51.3) for the Llama language decoder, and are trained using NVIDIA H100 GPUs.

## Analysis

### Evaluation metrics

Model performance was assessed on two complementary tasks: scene captioning and question-answering. For scene captioning from neural data, we evaluate the correspondence between model-generated captions and stimulus images using CLIPScore[35]. The noise ceiling is defined by the mean CLIPScore of MS COCO human captions[18]. To facilitate comparison with prior work, we also report standard captioning metrics, including BLEU, CIDEr, METEOR, and ROUGE, computed with the *pycocoevalcap* package[36] in table S1. For evaluating QA performance, generated and ground-truth answers were embedded using Qwen3-Embedding-8B[20], a state-of-the art language sentence embedder with high sensitivity to semantic nuances. Mirroring CLIPScore, we quantify performance as cosine similarity between embeddings of generated and ground-truth captions, to capture semantic correspondence despite potential syntactic variation. For both image captioning and question-answering, lower bounds are obtained by shuffling generated and ground-truth answers as well as evaluating the control model.

## Zero-shot training and evaluation

To test category level generalization we retrain CorText on reduced training datasets where all samples belonging to one COCO category are removed and evaluate the resulting models on test trials of the held-out category. We perform the analysis for three reduced datasets for subject 1, excluding the categories 'zebra', 'surfer', and 'airplane', respectively. Category trials are filtered by matching the following terms in the COCO captions: *zebra/zebras, surf/surfer/surfers/surfing/surfboard, airplane/airplanes/plane/planes*. One instance of CorText is trained for each of these reduced datasets.

Ground-truth COCO captions of the full training set for subject 1 are embedded using CLIP, and the generated captions for held-out categories of each model are embedded in the same space. Pairwise cosine distances are computed between embeddings, which we visualize with UMAP[37]. In addition, we prompt each model with a forced-choice question for all 3 held-out category test sets ("*What is in this image? Answer with one noun, chosen from [zebra, airplane, surfer]*" and record the models' accuracy.

## In-silico microstimulation

Face-selective and car-selective regions were identified using the functional localizers provided by the Natural Scenes Dataset, which contrasts responses to faces (adult and child) or cars against non-face / non-car stimuli using a two-sample t-test per surface vertex. We map these t-values to fsaverage space, and construct functional masks by thresholding at the top 1% of t-values, corresponding to 3,277 vertices. We select 20 trials without, and 20 trials with people or vehicles, and multiply the t-values contained in the functional mask by varying strengths, ranging from -5 to +5. These are then added to the original trial betas to either induce or reduce the face response (see Equation 1).

$$(1)\ Betas_{stimulated} = betas_{trial} + \beta \cdot mask$$

Using these modified neural data, we run inference for the 4 main trained instances of CorText. For each perturbed trial, we queried CorText with neutral prompts about scene content ("*What is in this image?*") using beam size = 1. For the people microstimulation experiment, we calculate the change in evidence by extracting the probabilities for the following person-related tokens were extracted: *person, people, man, woman, men, women, boy, girl*. In the vehicle microstimulation experiment, we do the same for the following tokens: *car, vehicle, bus, car, train, truck, plane, motor.* For each trial, we sum the softmax probabilities of these tokens and computed the log probability. Shifts in log-probabilities across microstimulation strengths relative to baseline quantified the effect of perturbation. In addition, we analyzed the proportion of generated captions explicitly mentioning the category of interest as a function of microstimulation strength. As control analysis, we randomly shuffle the position order of the localizer masks, and stimulate the neural data accordingly.

**Acknowledgments:**

We would like to thank Dirk Gütlin and Sean Gies for preliminary work guiding later developments of this project. In addition, we thank Philip Sulewski for the project name.

**Funding:**

ERC StG grant 101039524 TIME (VB, AD, TCK, DA), SNF grant n.203018 (AD), RTG GRK2340 DFG (DA). Compute resources for this project are in part funded by the Deutsche Forschungsgemeinschaft (DFG, German Research Foundation), project number 456666331.

**Author contributions:**

Conceptualization: VB, DA, AD, ST, TCK, PK
Methodology: VB, DA, TCK
Investigation: VB
Software: VB
Supervision: TCK
Writing – original draft: VB, TCK
Writing – review & editing: VB, TCK, AD, DA, ST, PK

**Competing interests:**

The authors declare no competing interests.

**Data availability:**

Datasets and language decoders used in this work are all available open-source. Custom models and code will be released upon publication.

**Materials & Correspondence:**

Correspondence and material requests should be addressed to VB and TCK.

# Supplementary Information

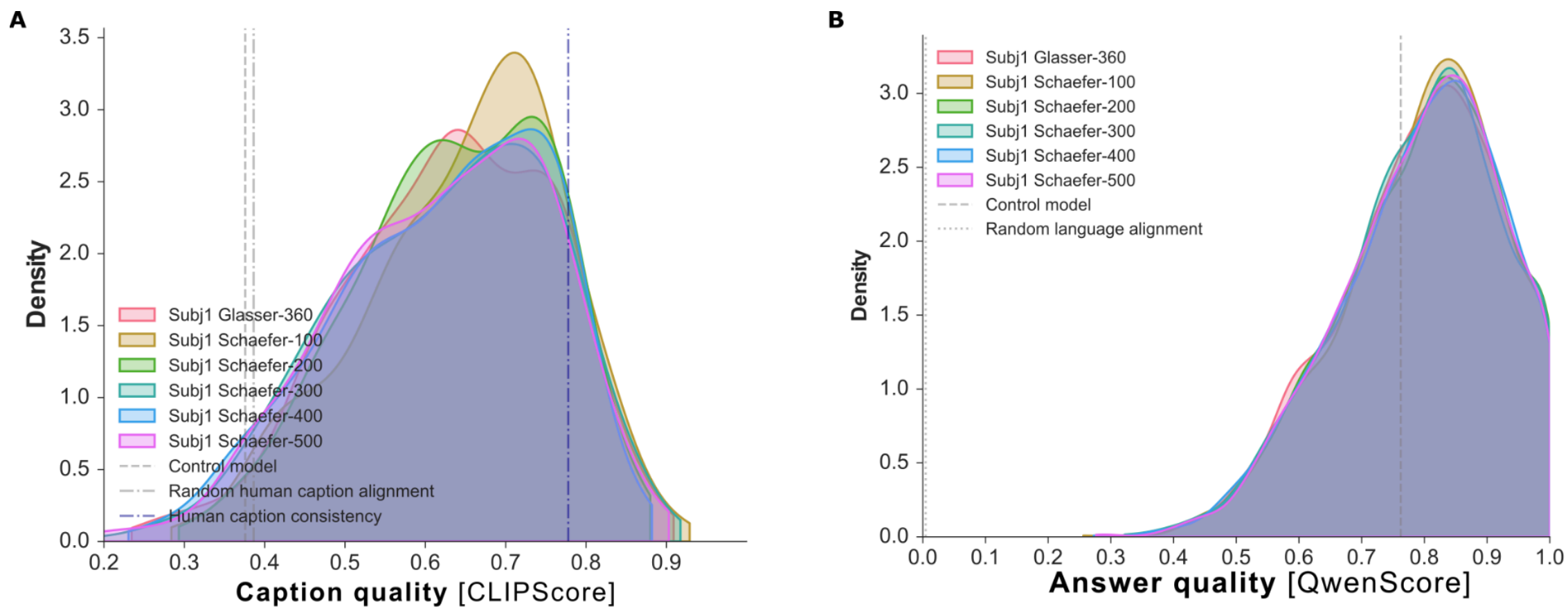

**Fig. S1. Performance of CorText using various cortical parcellations.** To determine how much the parcellation impacts performance ('all parcellations are wrong, but some are useful'), we also train CorText instantiations using different sizes of the Schaefer parcellation and the Glasser parcellation[32] for the brain tokenizer.

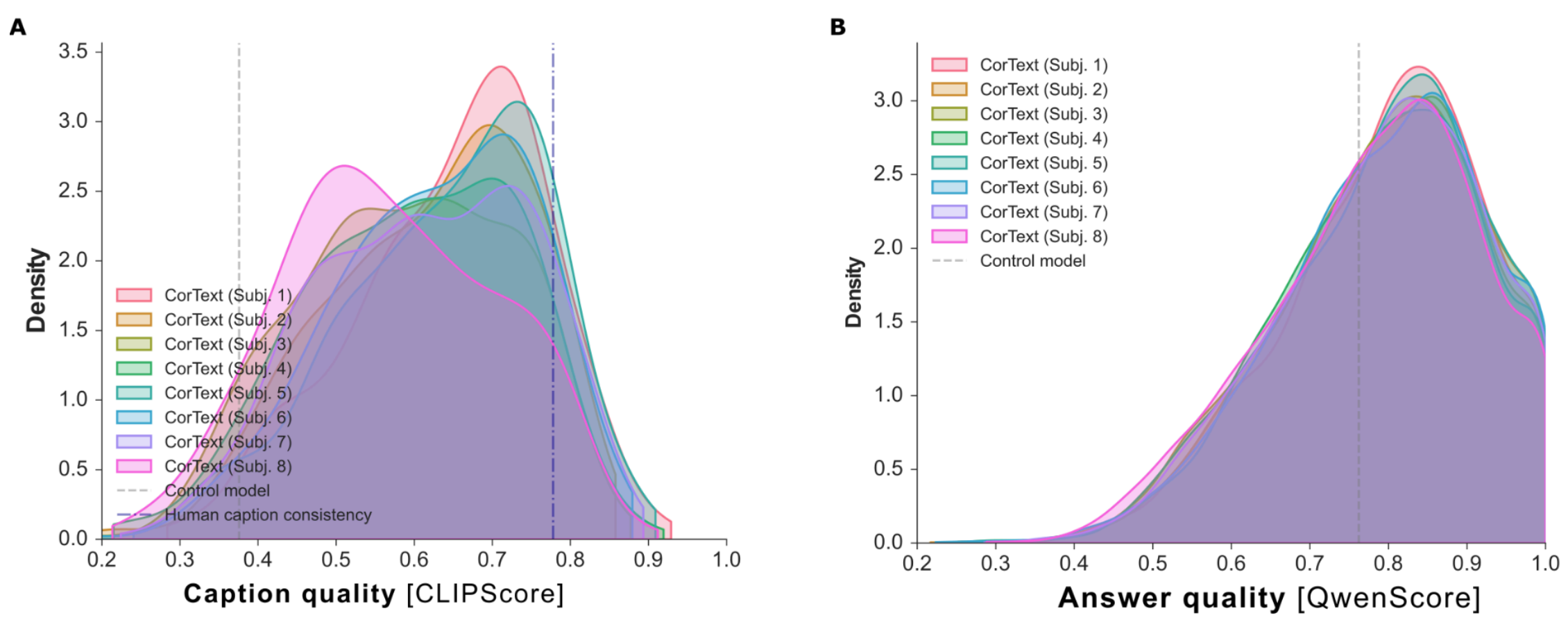

**Fig. S2. Performance of CorText models for all NSD subjects on the test set for each subject.**

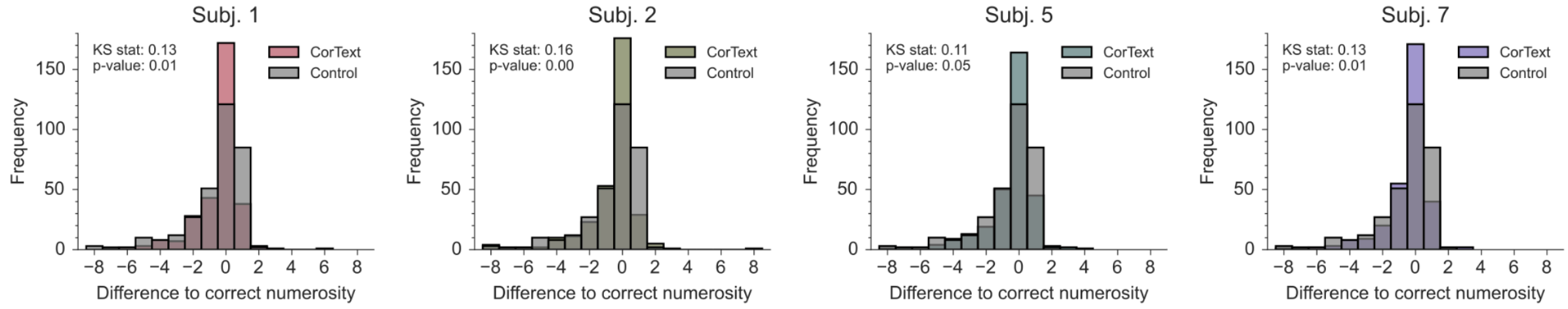

**Fig S3.** Accuracy of answers to questions about numerosity of objects in visual scenes in the test set for all main subject CorText models, versus the control model. All significant ($p<0.05$), Kolmogorov-Smirnov test between CorText and control distribution of predictions.

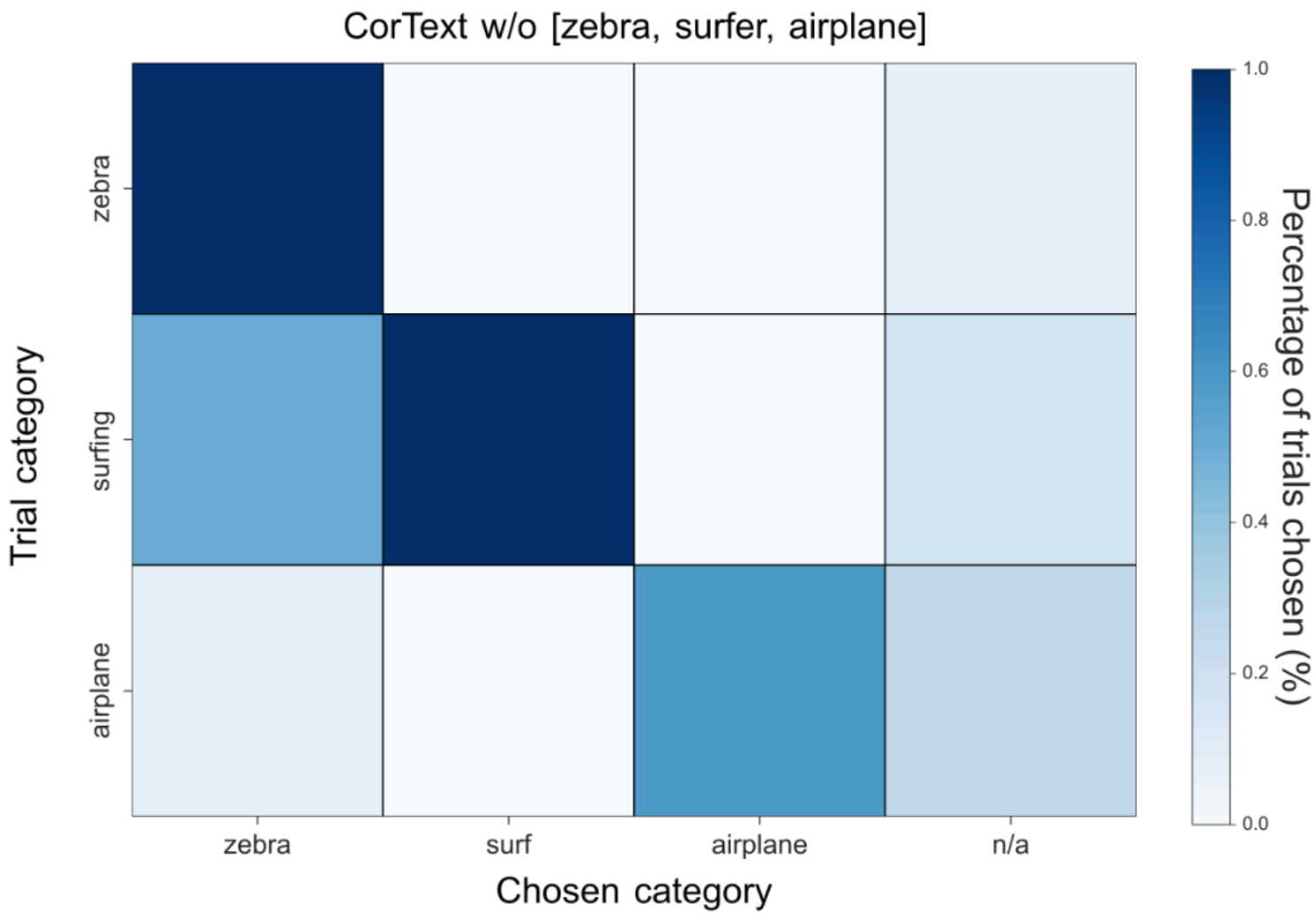

**Fig S4.** Testing the performance of CorText model trained without the zebra, surfer and airplane clusters in a choice experiment confirms generalization of the model to unseen semantic clusters.

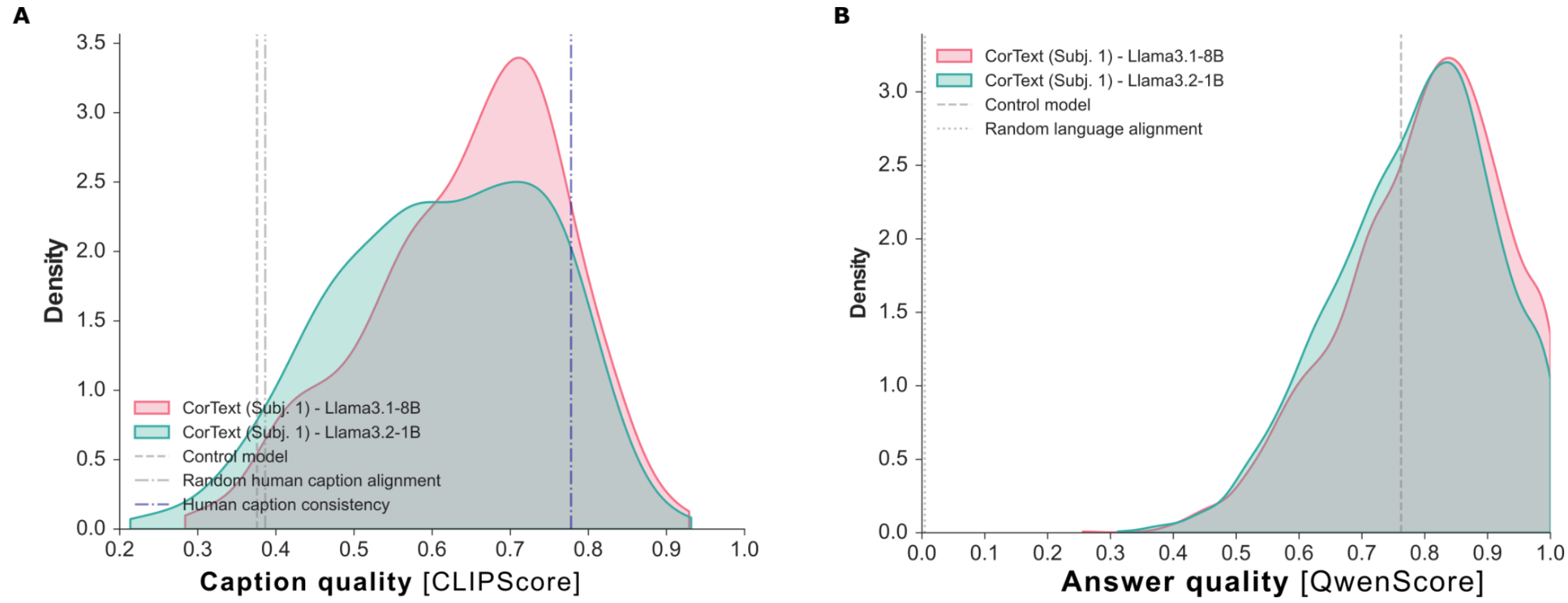

**Fig S5. Comparison of performance of CorText models under different language backbones.** Training a CorText model (Subject 1, Schaefer-100 parcellation) using Llama3.2-1B as language decoder does not perform as well as CorText based on Llama3.1-8B on decoding tasks on the test set.

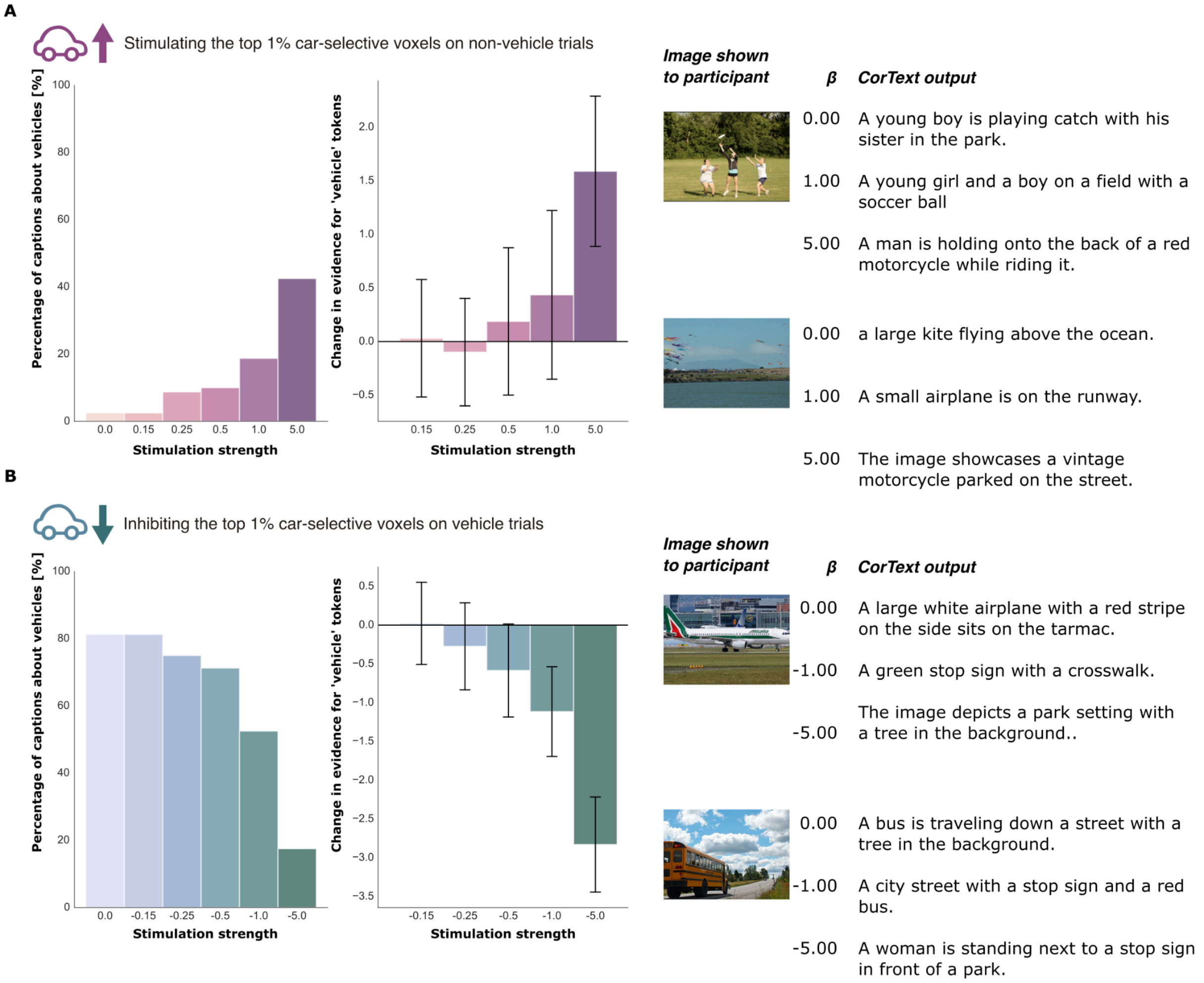

**Fig S6. Microstimulation experiment on car-selective vertices. (A)** Microstimulation of top 1% of car-selective vertices in non-vehicle trials increases the percentage of captions mentioning vehicle (left), as well as the evidence for 'vehicle' tokens (middle). **(B)** Inhibiting the top 1% of vertices with strong responses to the presence of cars in trials showing vehicles decreases the percentage of captions mentioning vehicles (left), as well as the evidence for 'vehicle' tokens (middle). Example captions (right) illustrate gradual semantic shifts induced by perturbation.

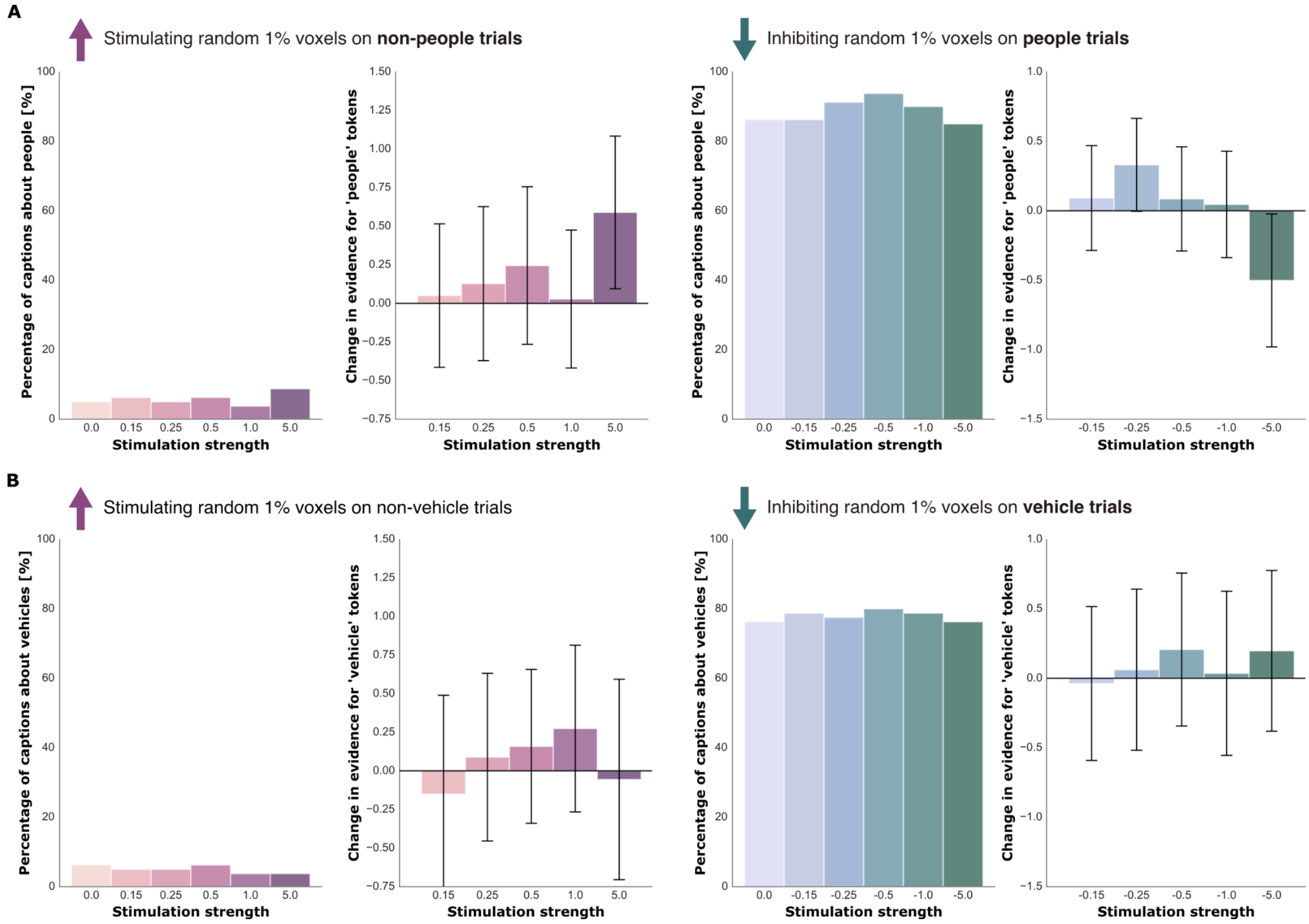

**Fig S7. Control microstimulation experiment using shuffled masks. (A)** Shuffling the mask of top 1% face-selective vertices and using it for stimulation on non-people trials and inhibition on people trials does not increase or respectively decrease the occurrence of people-related content in CorText's predictions. **(B)** Shuffling the mask of top 1% car-selective vertices and using it for stimulation on non-vehicle trials and inhibition on vehicle trials does not increase or respectively decrease the occurrence of vehicle-related content in CorText's predictions.

| Model | Bleu-1 | Bleu-2 | Bleu-3 | Bleu-4 | METEOR | ROUGE-L | CIDEr | CLIPScore | RefCLIPScore |
|---|---|---|---|---|---|---|---|---|---|
| **CorText (Subj. 1)** | 0.621 | 0.437 | 0.298 | 0.204 | 0.204 | 0.46 | 0.659 | 0.647 | 0.713 |
| **CorText (Subj. 2)** | 0.59 | 0.399 | 0.264 | 0.176 | 0.187 | 0.445 | 0.56 | 0.62 | 0.686 |
| **CorText (Subj. 3)** | 0.574 | 0.376 | 0.245 | 0.161 | 0.173 | 0.428 | 0.479 | 0.59 | 0.657 |
| **CorText (Subj. 4)** | 0.583 | 0.388 | 0.255 | 0.171 | 0.179 | 0.435 | 0.514 | 0.599 | 0.668 |
| **CorText (Subj. 5)** | 0.617 | 0.431 | 0.296 | 0.205 | 0.201 | 0.458 | 0.662 | 0.642 | 0.709 |
| **CorText (Subj. 6)** | 0.608 | 0.422 | 0.288 | 0.198 | 0.194 | 0.454 | 0.595 | 0.627 | 0.691 |
| **CorText (Subj. 7)** | 0.582 | 0.39 | 0.253 | 0.166 | 0.184 | 0.44 | 0.549 | 0.615 | 0.682 |
| **CorText (Subj. 8)** | 0.55 | 0.356 | 0.224 | 0.14 | 0.164 | 0.411 | 0.412 | 0.569 | 0.638 |
| **Control** | 0.39 | 0.192 | 0.89 | 0.48 | 0.108 | 0.307 | 0.63 | 0.376 | 0.446 |

**Table S1. Captioning performance across subjects.** Quantitative evaluation of CorText answers in the captioning task on the test set, using standard syntactic and semantic metrics (BLEU, METEOR, ROUGE, CIDEr, CLIPScore).

| Question | CorText answer | Human caption |
|---|---|---|
| **Describe the important features of the scene in this image:** | A man is making a pizza in a kitchen. | a couple of people are cooking in a room |
| **What is in this picture?** | A man riding a surfboard in the ocean. | A person in a wetsuit surfing on a turquoise wave. |
| **Give a concise and descriptive caption of this image:** | A person riding a surfboard in the ocean. | a windsurfer some water a hill sand and some kayaks |
| **Give a concise and descriptive caption of this image:** | A building with a clock on the side of it. | A brick building with a clock on it and pathway. |
| **Provide a detailed description of the given image in one sentence:** | A plate of food with broccoli carrots and rice. | a plate of yummy food of some kind |
| **Give a concise and descriptive caption of this image:** | A plate with a sandwich and a cup of coffee on it. | A cup of coffee on a plate with a spoon. |
| **Describe the following image in detail in one sentence:** | A group of kids playing with frisbees on the beach. | Three young women are trying to catch a frisbee. |
| **Describe the following image in detail in one sentence:** | A brown and white bear sitting on a rock. | a cat is sitting on top of a vehicle |
| **What is in this picture?** | A surfer riding a wave on a surfboard. | A man in swim shorts is riding a surf board. |
| **What is in this picture?** | A commercial airliner is flying in the sky. | A lone airplane is flying high against a grey sky. |
| **Provide a detailed description of the given image in one sentence:** | A young boy riding a surfboard in the ocean. | A young child riding a wave on top of a board. |
| **Describe the important features of the scene in this image:** | A group of sheep grazing in a field. | A bear that is standing in the water. |
| **Provide a detailed description of the given image in one sentence:** | A man is sitting on a bed with a laptop on his lap. | Man laying on top of bed in green towel working on laptop. |
| **Describe the following image in detail in one sentence:** | A train is traveling on a bridge over a body of water. | A train traveling over a river on a bridge. |
| **What is in this picture?** | A living room with couches chairs and a fireplace. | A living room that has a couch in it. |

| **Describe the following image in detail in one sentence:** | A man wearing a suit and tie is holding a cell phone. | a man with a brown suit  sitting at a table with microphone  listening. |
|---|---|---|
| **Give a concise and descriptive caption of this image:** | A group of cows grazing on a grassy field near a body of water. | A herd of cattle grazing on a grassy hill. |
| **What is in this picture?** | A man in a white shirt and blue pants is holding a white baseball bat. | A young man holding a ball in front of a soccer net. |
| **Provide a detailed description of the given image in one sentence:** | A bathroom with a toilet  sink and shower. | A very small bathroom stall with a toilet and several rolls of toilet paper. |
| **Describe the following image in detail in one sentence:** | A young boy is sitting on a bed with a large surfboard next to him. | A toddler reading a book to a baby. |
| **What is in this picture?** | A group of people standing in front of a fire truck. | A view of a bunch of people skiing and posing for a picture. |
| **Describe the important features of the scene in this image:** | A clock tower with a clock face on it. | A very tall clock tower with two clocks on it. |
| **Give a concise and descriptive caption of this image:** | A group of people walking through a field of tall grass. | Two cowboys on horseback are riding along a dirt road. |
| **Give a concise and descriptive caption of this image:** | A white and black horse standing on a rocky hill. | A giraffe standing on top of a lush green field. |
| **Give a concise and descriptive caption of this image:** | A man and a woman sitting on a couch with a dog. | Two brown teddy bears sitting on floor next to a piano. |
| **Give a concise and descriptive caption of this image:** | A bowl of apples and a glass of orange juice on a table. | An apple between two pomegranates siting on a table covered by a white table cloth. |
| **Describe the following image in detail in one sentence:** | A small airplane is flying over the water. | A small propeller plane is at a sandy beach. |
| **Provide a detailed description of the given image in one sentence:** | Two birds standing on top of a pile of dirt. | a couple of birds sit on a branch with each other |
| **Describe the following image in detail in one sentence:** | A young girl sitting on a bed in a hotel room. | A little girl sitting on top of a bed near pillows. |
| **Describe the following image in detail in one sentence:** | A piece of banana on a plate with a fork next to it. | Glass of frothy white liquid sitting on wooden surface next to a banana |

**Table S2. Samples of CorText decoding results on the scene captioning task (subject 1, test set).**

| Question | CorText answer | Human caption |
|---|---|---|
| **What is in this picture?** | A group of people are preparing food in a kitchen. | a couple of people are cooking in a room |
| **Describe the important features of the scene in this image:** | A person riding a skateboard on a concrete ramp. | A person in a wetsuit surfing on a turquoise wave. |
| **Provide a detailed description of the given image in one sentence:** | A man is riding a snowboard on a snow covered beach. | a windsurfer some water a hill sand and some kayaks |
| **Give a concise and descriptive caption of this image:** | A black and white photo of a street with a building in the background. | A brick building with a clock on it and pathway. |
| **Describe the important features of the scene in this image:** | A plate of food on a table with a napkin. | a plate of yummy food of some kind |
| **Give a concise and descriptive caption of this image:** | A plate of food on a table next to a bowl of fruit. | A cup of coffee on a plate with a spoon. |
| **Provide a detailed description of the given image in one sentence:** | A group of people are playing a game of frisbee. | Three young women are trying to catch a frisbee. |
| **What is in this picture?** | A black and white dog laying on a park bench. | a cat is sitting on top of a vehicle |
| **What is in this picture?** | A person riding a wave on a surfboard. | A man in swim shorts is riding a surf board. |
| **Describe the following image in detail in one sentence:** | A blue and white airplane sitting on the tarmac. | A lone airplane is flying high against a grey sky. |
| **Describe the important features of the scene in this image:** | A man standing on a surfboard in the ocean. | A young child riding a wave on top of a board. |
| **Describe the following image in detail in one sentence:** | A bird standing on a grassy field with a bucket next to it. | A bear that is standing in the water. |
| **Describe the following image in detail in one sentence:** | A woman laying on a bed with a cat. | Man laying on top of bed in green towel working on laptop. |
| **Give a concise and descriptive caption of this image:** | A train on a track with a building in the background. | A train traveling over a river on a bridge. |

| **Give a concise and descriptive caption of this image:** | A living room with a desk and a laptop. | A living room that has a couch in it. |
|---|---|---|
| **What is in this picture?** | A man sitting in front of a computer. | a man with a brown suit sitting at a table with microphone listening. |
| **Describe the following image in detail in one sentence:** | A flock of geese flying over a field. | A herd of cattle grazing on a grassy hill. |
| **What is in this picture?** | A young girl standing on a baseball field holding a bat. | A young man holding a ball in front of a soccer net. |
| **Describe the important features of the scene in this image:** | A bathroom with a toilet sink and shower. | A very small bathroom stall with a toilet and several rolls of toilet paper. |
| **Give a concise and descriptive caption of this image:** | A young girl holding a baby while smiling at the camera. | A toddler reading a book to a baby. |
| **What is in this picture?** | A group of people posing for a photo in front of an airplane. | A view of a bunch of people skiing and posing for a picture. |
| **Describe the important features of the scene in this image:** | A clock tower with a large clock on it. | A very tall clock tower with two clocks on it. |
| **Describe the following image in detail in one sentence:** | A man and a woman standing next to each other on a grassy field. | Two cowboys on horseback are riding along a dirt road. |
| **What is in this picture?** | A giraffe standing in a field with trees in the background. | A giraffe standing on top of a lush green field. |
| **Provide a detailed description of the given image in one sentence:** | A black and white photo of a stuffed animal sitting on a table. | Two brown teddy bears sitting on floor next to a piano. |
| **Give a concise and descriptive caption of this image:** | A close up of a plate with two oranges on it. | An apple between two pomegranates siting on a table covered by a white table cloth. |
| **Describe the important features of the scene in this image:** | A red fire truck is parked on a dirt road. | A small propeller plane is at a sandy beach. |
| **Provide a detailed description of the given image in one sentence:** | Two white birds sitting on top of a large rock. | a couple of birds sit on a branch with each other |
| **Describe the following image in detail in one sentence:** | A young girl sitting on a couch with a teddy bear. | A little girl sitting on top of a bed near pillows. |

| **Give a concise and descriptive caption of this image:** | A plate with a slice of cake and a fork. | Glass of frothy white liquid sitting on wooden surface next to a banana |
|---|---|---|

**Table S3. Samples of CorText decoding results on the scene captioning task (subject 2, test set).**

| Question | CorText answer | Human caption |
|---|---|---|
| **Describe the important features of the scene in this image:** | A woman is cooking a meal on a grill. | a couple of people are cooking in a room |
| **Provide a detailed description of the given image in one sentence:** | A person riding a surfboard on a wave. | A person in a wetsuit surfing on a turquoise wave. |
| **Give a concise and descriptive caption of this image:** | A person riding a surfboard in the ocean. | a windsurfer some water a hill sand and some kayaks |
| **What is in this picture?** | A bunch of wooden boats are lined up on a dock. | A brick building with a clock on it and pathway. |
| **Describe the following image in detail in one sentence:** | A plate of food on a table with a napkin. | a plate of yummy food of some kind |
| **Provide a detailed description of the given image in one sentence:** | A table with a variety of food and drinks on it. | A cup of coffee on a plate with a spoon. |
| **Give a concise and descriptive caption of this image:** | A group of young men playing soccer on a field. | Three young women are trying to catch a frisbee. |
| **Provide a detailed description of the given image in one sentence:** | A small dog sitting on top of a large rock. | a cat is sitting on top of a vehicle |
| **Provide a detailed description of the given image in one sentence:** | A man riding a skateboard on a ramp. | A man in swim shorts is riding a surf board. |
| **Describe the important features of the scene in this image:** | A large airplane is parked on a runway. | A lone airplane is flying high against a grey sky. |
| **Give a concise and descriptive caption of this image:** | A man riding a surfboard in the ocean. | A young child riding a wave on top of a board. |
| **Describe the following image in detail in one sentence:** | A white dog standing in a field with a frisbee in its mouth. | A bear that is standing in the water. |
| **Provide a detailed description of the given image in one sentence:** | A woman sitting on a bed with a teddy bear. | Man laying on top of bed in green towel working on laptop. |
| **Describe the important features of the scene in this image:** | A train is traveling down a track near a body of water. | A train traveling over a river on a bridge. |
| **Provide a detailed description of the given image in one sentence:** | A living room with a couch and a coffee table. | A living room that has a couch in it. |
| **Describe the following image in detail in one sentence:** | A man in a suit and tie is standing in front of a tree. | a man with a brown suit sitting at a table with microphone listening. |
| **What is in this picture?** | A herd of cows grazing in a field. | A herd of cattle grazing on a grassy hill. |
| **Describe the following image in detail in one sentence:** | A man in a yellow shirt holding a tennis racket. | A young man holding a ball in front of a soccer net. |
| **Give a concise and descriptive caption of this image:** | A white toilet in a white bathroom with a white sink. | A very small bathroom stall with a toilet and several rolls of toilet paper. |
| **Provide a detailed description of the given image in one sentence:** | A young girl laying on a bed with a stuffed animal next to her. | A toddler reading a book to a baby. |
| **Describe the following image in detail in one sentence:** | A group of people posing for a photo in a field. | A view of a bunch of people skiing and posing for a picture. |
| **Describe the important features of the scene in this image:** | A clock tower with a large clock face on it. | A very tall clock tower with two clocks on it. |
| **Give a concise and descriptive caption of this image:** | A group of people standing on a grassy field. | Two cowboys on horseback are riding along a dirt road. |
| **What is in this picture?** | A giraffe is standing in a field with trees in the background. | A giraffe standing on top of a lush green field. |
| **Describe the important features of the scene in this image:** | A white dog sitting on top of a wooden bench. | Two brown teddy bears sitting on floor next to a piano. |
| **Describe the following image in detail in one sentence:** | A bunch of oranges sitting on top of a table. | An apple between two pomegranates siting on a table covered by a white table cloth. |
| **Give a concise and descriptive caption of this image:** | A red and white train on a snowy track. | A small propeller plane is at a sandy beach. |
| **What is in this picture?** | Two large birds standing on a rocky outcropping near a body of water. | a couple of birds sit on a branch with each other |

| Describe the following image in detail in one sentence: | A young woman sitting on a bed with a blanket. | A little girl sitting on top of a bed near pillows. |
|---|---|---|
| **Describe the following image in detail in one sentence:** | A glass of orange juice with a slice of orange in it. | Glass of frothy white liquid sitting on wooden surface next to a banana |

**Table S4. Samples of CorText decoding results on the scene captioning task (subject 5, test set).**

| **Question** | **CorText answer** | **Human caption** |
|---|---|---|
| **Give a concise and descriptive caption of this image:** | A man in a kitchen preparing a meal in a stainless steel pot. | a couple of people are cooking in a room |
| **Provide a detailed description of the given image in one sentence:** | A surfer riding a wave in the ocean. | A person in a wetsuit surfing on a turquoise wave. |
| **Give a concise and descriptive caption of this image:** | A group of people flying kites on a beach. | a windsurfer some water a hill sand and some kayaks |
| **Give a concise and descriptive caption of this image:** | A red and white train sitting on the tracks. | A brick building with a clock on it and pathway. |
| **Provide a detailed description of the given image in one sentence:** | A plate of food with a variety of toppings. | a plate of yummy food of some kind |
| **Describe the important features of the scene in this image:** | A plate of food on a table next to a soda. | A cup of coffee on a plate with a spoon. |
| **Give a concise and descriptive caption of this image:** | A man and a woman playing a game of tennis. | Three young women are trying to catch a frisbee. |
| **Describe the following image in detail in one sentence:** | A black and white photo of a cat sitting on a car seat. | a cat is sitting on top of a vehicle |
| **Give a concise and descriptive caption of this image:** | A person in a red jacket is skiing down a snow covered hill. | A man in swim shorts is riding a surf board. |
| **Give a concise and descriptive caption of this image:** | A white and red airplane flying in the sky. | A lone airplane is flying high against a grey sky. |
| **Describe the important features of the scene in this image:** | A young boy is riding a surfboard in the ocean. | A young child riding a wave on top of a board. |
| **Describe the following image in detail in one sentence:** | A bear is standing in a stream of water. | A bear that is standing in the water. |
| **What is in this picture?** | A woman is sitting on a couch with a baby in her arms. | Man laying on top of bed in green towel working on laptop. |
| **Give a concise and descriptive caption of this image:** | A large boat is docked in the water. | A train traveling over a river on a bridge. |
| **Provide a detailed description of the given image in one sentence:** | A bunch of cars parked in a parking lot next to a train station. | A living room that has a couch in it. |

| **Give a concise and descriptive caption of this image:** | A woman smiling and holding a baby in a stroller. | a man with a brown suit sitting at a table with microphone listening. |
|---|---|---|
| **Provide a detailed description of the given image in one sentence:** | A herd of zebras grazing in a field. | A herd of cattle grazing on a grassy hill. |
| **Describe the important features of the scene in this image:** | A woman in a white tank top playing tennis. | A young man holding a ball in front of a soccer net. |
| **Provide a detailed description of the given image in one sentence:** | A bathroom with a toilet sink and mirror. | A very small bathroom stall with a toilet and several rolls of toilet paper. |
| **Describe the following image in detail in one sentence:** | A man and a woman sitting on a bench with a dog. | A toddler reading a book to a baby. |
| **Describe the important features of the scene in this image:** | A group of people standing in front of a building. | A view of a bunch of people skiing and posing for a picture. |
| **Provide a detailed description of the given image in one sentence:** | A clock tower with a clock on the top. | A very tall clock tower with two clocks on it. |
| **Describe the important features of the scene in this image:** | A group of people riding horses down a dirt road. | Two cowboys on horseback are riding along a dirt road. |
| **Describe the important features of the scene in this image:** | A giraffe standing in a grassy field. | A giraffe standing on top of a lush green field. |
| **Give a concise and descriptive caption of this image:** | A teddy bear sitting on a bench in a park. | Two brown teddy bears sitting on floor next to a piano. |
| **Describe the important features of the scene in this image:** | A bowl of fruit sitting on top of a table. | An apple between two pomegranates siting on a table covered by a white table cloth. |
| **Provide a detailed description of the given image in one sentence:** | A small airplane sitting on a grassy field. | A small propeller plane is at a sandy beach. |
| **Describe the following image in detail in one sentence:** | A group of elephants standing in a field with flowers. | a couple of birds sit on a branch with each other |
| **Describe the following image in detail in one sentence:** | A young boy sitting on a couch with a teddy bear. | A little girl sitting on top of a bed near pillows. |
| **Give a concise and descriptive caption of this image:** | A slice of pizza on a plate with a fork next to it. | Glass of frothy white liquid sitting on wooden surface next to a banana |

**Table S5. Samples of CorText decoding results on the scene captioning task (subject 7, test set).**

| Question | CorText answer | Human caption |
|---|---|---|
| **Can you tell me more about the environment they are in?** | Certainly. The environment appears to be an indoor setting  possibly a kitchen or a dining area. There's a window in the background  and the | They are in a well-equipped commercial kitchen with stainless steel appliances and surfaces. There are various kitchen tools and utensils around  such as pots |
| **Where is the sun located in the image?** | The sun is located in the upper right corner of the image  casting a warm glow over the scene. | The exact location of the sun isn't visible in the image  but from the lighting  it seems to be shining from above and possibly slightly from |
| **How many canoes can you spot in the image?** | I can spot two canoes in the image. | I can see a total of five canoes in the image. |
| **Is there any prominent feature on the church's tower?** | Yes  the church's tower has a prominent clock on it. | Yes  the church's tower has a clock visible on it. |
| **Do the potatoes look like they're fried or boiled?** | Based on the visual appearance  the potatoes seem to be boiled or steamed. They have a smooth texture and appear to be cooked through. However | The potatoes have a slightly golden appearance  which might suggest they've been fried or sautéed. They don't have the pale appearance typical |
| **Can you elaborate on the latte art design on the coffee?** | Certainly! The latte art design on the coffee appears to be a stylized image of a heart. It's a common design found on l | Yes  the latte art design on the coffee appears to be a classic rosette or tulip pattern. This design is achieved by pouring |
| **What could be a potential backstory for this scene based on what you see?** | Based on the image  it seems like a casual day at the beach for these individuals. They might have come to the beach to enjoy the sun | Based on the image  it appears to be a casual day at the park. The three individuals might be friends or teammates practicing or simply enjoying a |
| **What might have caused the cat to sit on the car?** | While I can't determine the exact reason  cats are known to be curious and often choose to sit in unexpected places. It could be due to | While I can't know for sure  cats often seek elevated positions for a variety of reasons. Sitting on a car can give the cat a better |
| **What color is the water?** | The water is blue. | The water appears to be a deep shade of blue. |
| **Based on the landing gears being extended  what can you infer about the airplane's current phase of flight?** | Given that the landing gears are extended  it suggests that the airplane is in the final phase of its flight  preparing to land. This is a | When an airplane has its landing gears extended  it typically indicates that the airplane is in the final phases of its flight  preparing to land. Given |

| Can you describe any other significant landmarks or objects in the picture? | Apart from the person and the surfboard there are some palm trees in the background which are common in coastal areas. The image also captures | Beyond the surfer and the wave the image predominantly features the open sea. There are no specific landmarks or significant other objects visible. |
|---|---|---|
| **What can you see in the image?** | I see a group of sheep grazing in a grassy field. They appear to be in a natural setting possibly a meadow or pasture. | In the image I can see a bear in the water surrounded by several logs. The bear appears to be mid-way in the water with |
| **Is there any jewelry visible in the image?** | Yes there is a necklace visible on the woman's neck. | Yes there appears to be a bracelet on the person's wrist. |
| **Can you explain the importance of bullet trains?** | Certainly! Bullet trains also known as high-speed trains are designed to travel at extremely high speeds often exceeding 300 kilometers per hour. | Absolutely. Bullet trains also known as high-speed trains play a crucial role in modern transportation systems. They offer several advantages: |
| **What's the relationship between the candle and the igniter in terms of positioning?** | The candle and the igniter are positioned close to each other with the igniter being slightly to the right of the candle. The igniter | The igniter is positioned to the right of the white candle on the coffee table. |
| **Can you describe the main subject in the image?** | Certainly. The main subject in the image is a man wearing a suit and tie. He appears to be looking directly at the camera with an expression | The main subject in the image is a man possibly in a professional setting sitting behind a desk or podium. He appears to be in the |
| **Besides the bulls are there any other objects or features in the foreground?** | Yes in the foreground there are some bushes and what appears to be a fence or a boundary. | Yes there are several logs scattered around the field where the bulls are. |
| **What is the position of the soccer ball in relation to the goalkeeper?** | The soccer ball is in front of the goalkeeper suggesting that the goalkeeper is in the process of catching or attempting to catch the ball. | The soccer ball is positioned in front of the goalkeeper indicating that it might be approaching him or that he has just saved it. The ball is |
| **How would you describe the design aesthetics of this bathroom based on the image?** | The bathroom has a minimalist design with a focus on functionality. The walls are painted a light color which gives the space an airy feel. The | The bathroom showcases a minimalist and modern design aesthetic. The choice of neutral colors like light browns and beige gives it a clean and sophisticated look. |
| **Is there any interaction between the two children?** | Yes the child on the left appears to be holding the surfboard while the child on the right seems to be looking at the surfboard | Yes there seems to be an interaction between the two children. The younger child is closely looking at the book while the older child is holding it |
| **Based on the attire and equipment what activity do you think these people were** | Based on the attire and equipment they're wearing it appears they were preparing for or participating in a | Based on their attire and equipment it seems that these people were involved in or preparing for skiing or snowboarding. |

| **involved in or preparing for?** | skiing or snowboarding event. The presence of | |
|---|---|---|
| **Can you describe the condition of the tower? Does it look new or old?** | The tower appears to be old  with visible signs of wear and tear. It has a weathered appearance  suggesting that it has been exposed to | The tower looks relatively old  with signs of wear and aging visible on its surface. It appears to be well-maintained  but its design |
| **What are the activities these people are involved in?** | The people appear to be walking in a forested area. One of them is carrying a backpack  suggesting they might be on a hike or excursion | Both people are riding horses. |
| **Is there any water source visible in the image?** | No  there is no visible water source in the image. | No  there is no visible water source in the image. |
| **How many teddy bears are there?** | There are two teddy bears in the image. | There are two teddy bears in the image. |
| **What is the position of the green apple relative to the red apples?** | The green apple is positioned to the left of the red apples. | The green apple is positioned between the two red apples. |
| **What is the condition of the weather in the background?** | The weather in the background appears to be overcast with clouds suggesting it might be a cloudy day. | The weather appears to be partly cloudy with blue skies and some scattered clouds. |
| **Is there any interaction or relationship you can infer between the two birds from the image?** | Based on the image  the two birds appear to be in close proximity to each other  suggesting they might be interacting  or have a relationship. However | The two birds seem to be in close proximity and facing each other. This could indicate some form of interaction or communication between them. They might be |
| **What can you see in the image?** | I see a young girl sitting on a bed. She's wearing a dress and has a backpack on her back. There's a window behind her | I see a young girl sitting on a bed in what appears to be a hotel room. She's wearing a yellow shirt and is sitting cross-legged |
| **What objects can you see in the image?** | In the image  I can see a banana  a fork  and a plate. | In the image  I can see a banana and a glass containing a drink or some kind of liquid with specks in it. |

**Table S6. Samples of CorText decoding results on the question-answering task (subject 1, test set).**

| Question | CorText answer | Human caption |
|---|---|---|
| **How many chefs are there in the image?** | There are two chefs in the image. | There are two chefs in the image. |
| **Describe the environment around the surfer.** | The environment around the surfer consists of a sandy beach with a clear blue sky above. There are some palm trees visible in the background and | The surfer is in the ocean surrounded by waves. The waters have varying shades of blue and green and there's some frothy white |
| **How would you describe the relative position of the windsurfer and the canoes?** | The windsurfer is positioned closer to the shore while the canoes are further out in the water. | The windsurfer is positioned on the water a bit farther from the shore. The canoes are on the sandy shore with some of |
| **Are there any trees in the vicinity?** | No there are no trees visible in the image. | Yes there are trees surrounding the church especially towards the right side of the image. |
| **Based on the appearance can you guess the possible cuisine or dish it might be?** | Based on the visual content it appears to be a vegetarian dish possibly a pasta or noodle-based meal with various vegetables. The presence of | Based on the visual content and considering the combination of diced potatoes and green onions it could be a part of various dishes from different cuisines. |
| **What's the object located on the top right corner?** | The object located on the top right corner appears to be a spoon. | The object located on the top right corner is a glass of water. |
| **Are there any other objects or equipment around them related to their activity?** | Yes there are frisbees on the ground near the players indicating that they are actively playing the game. | I don't see any other equipment specifically related to their frisbee activity but there are some cones visible in the image which may be |
| **What is the color of the cat?** | The cat is black. | The cat has a mix of gray and white fur. |
| **Can you tell me more about the board the surfer is using?** | Certainly. The surfboard is a long narrow and flat piece of wood. It has a rounded nose and a pointed tail. The surface | Certainly! The surfer is using a surfboard that appears to be a shortboard judging by its shape and size. Shortboards are designed |
| **What colors are prominent on the airplane?** | The prominent colors on the airplane are white blue and red. | The airplane is predominantly white with blue accents. It also has a blue tail fin that features a red and blue logo. |
| **What is the color of the surfboard?** | The surfboard is blue. | The surfboard appears to be blue. |

| **Is the bear closer to the left or right edge of the image?** | The bear is closer to the left edge of the image. | The bear is closer to the left edge of the image. |
|---|---|---|
| **Are there any electronic devices besides the laptop?** | Yes there is a cellular telephone visible in the image. | Yes there is a telephone present in the image. |
| **If there was a boat in the river where would it be in relation to the train?** | If there was a boat in the river it would likely be in the foreground closer to the viewer and possibly near the riverbank. | If there was a boat in the river given the current positioning of the bullet train on the bridge the boat could be either directly below the |
| **What's the relationship between the candle and the igniter in terms of positioning?** | The candle is positioned to the right of the igniter. | The igniter is positioned to the right of the white candle on the coffee table. |
| **Are there any accessories he's wearing or has nearby?** | Yes he is wearing a watch on his wrist. Additionally there's a wallet on the table next to him. | Yes besides the spectacles he is wearing a wristwatch. There's also a microphone placed close to him indicating he might be addressing |
| **Besides the bulls are there any other objects or features in the foreground?** | Yes in the foreground there's a fence that runs parallel to the road. Beyond the fence there's a field with tall grass. | Yes there are several logs scattered around the field where the bulls are. |
| **Describe the weather or atmosphere based on the image.** | Based on the image it appears to be a sunny day with clear skies. There are no clouds visible in the sky suggesting a clear and | Based on the image the weather appears to be cloudy or overcast. The field looks slightly damp possibly suggesting that it has rained earlier |
| **How would you describe the design aesthetics of this bathroom based on the image?** | The bathroom has a minimalist design with a focus on clean lines and functionality. The white walls and floor provide a neutral background while the toilet and | The bathroom showcases a minimalist and modern design aesthetic. The choice of neutral colors like light browns and beige gives it a clean and sophisticated look. |
| **What is the title or theme of the book the older child is holding?** | The title of the book the older child is holding is "The Magic School Bus". | The book the older child is holding is titled "The Lady with the Alligator Purse " and it appears to have illustrations related to alligators |
| **What is the main theme of the photo?** | The main theme of the photo appears to be a group of people posing for a photograph possibly at a military or ceremonial event given the attire | The main theme of the photo is a group of people who seem to be at a ski resort during the nighttime. They are dressed in winter clothing |
| **What seems to be the architectural style of the tower?** | The tower appears to be in a Gothic Revival style characterized by its pointed arches ribbed vaults and the use of pointed | The architectural style of the tower appears to be neoclassical characterized by its symmetrical design classical columns and grand facades. |

| **Can you describe the horses in detail?** | Certainly. There are two horses in the image. The first horse is standing on the left side of the image. It has a brown coat with | Certainly. There are two horses: one is brown and the other is white with some grayish spots. Both horses appear to be in good |
|---|---|---|
| **Where is the giraffe located?** | The giraffe is located in the center of the image standing on a grassy area surrounded by trees. | The giraffe is standing in a grassy field with trees in the background. |
| **Is there any person playing the piano?** | No there is no person playing the piano in the image. | No there isn't anyone playing the piano in the image. |
| **How many apples are there and can you describe their colors?** | There are two apples in the image. One is red and the other is green. | There are three apples in the image. Two of them are red with a hint of yellow and one is green. |
| **Can you describe the surroundings of the airplane in more detail?** | Certainly. The airplane is situated on a grassy surface surrounded by trees. There are some buildings and structures in the background but they are | Certainly. The airplane is situated on a sandy beach with the ocean waves visible in the background. The beach appears to be expansive with the horizon |
| **Based on their features can you infer any specific type or species of these birds?** | The birds appear to be some species of waterfowl. Given their size plumage and beak shape they could be a type | The birds appear to be Bee-eaters identifiable by their colorful plumage and sleek body structure. Bee-eaters are known for their vibrant colors |
| **Where is the girl in relation to the lamp?** | The girl is positioned to the left of the lamp. | The girl is sitting on the bed and the lamp is to her side on a bedside table. The lamp is on her left side when looking |
| **Would the drink be considered thick or thin based on its appearance?** | Based on the image the drink appears to be thick possibly a smoothie or a milkshake given its creamy texture. | Based on its appearance the drink looks thick resembling the consistency of a smoothie or shake. |

**Table S7. Samples of CorText decoding results on the question-answering task (subject 2, test set).**

| Question | CorText answer | Human caption |
|---|---|---|
| **What are they wearing?** | The woman is wearing a white shirt and a hat  while the man is wearing a black shirt and a hat as well. | Both chefs are wearing white chef uniforms with hats. One of them has an apron with visible logos and text. |
| **Describe the environment around the surfer.** | The surfer is in an open ocean setting with a clear blue sky above. There are no visible landmasses in the immediate background. The | The surfer is in the ocean  surrounded by waves. The waters have varying shades of blue and green  and there's some frothy white |
| **How many canoes can you spot in the image?** | I can see two canoes in the image. | I can see a total of five canoes in the image. |
| **Is there any prominent feature on the church's tower?** | Yes  the church's tower has a clock on it. | Yes  the church's tower has a clock visible on it. |
| **What does the image mainly depict?** | The image primarily depicts a meal consisting of various dishes placed on a table. The main focus is on the pizza  which is topped with various ingredients | The image primarily showcases a plate of food. The main ingredients visible on the plate are diced potatoes and chopped green onions. |
| **Can you elaborate on the latte art design on the coffee?** | Certainly. The latte art design on the coffee features a floral pattern. The design is intricate  showcasing a mix of flowers and leaves. The | Yes  the latte art design on the coffee appears to be a classic rosette or tulip pattern. This design is achieved by pouring |
| **Describe the setting or environment where they are.** | The setting appears to be an outdoor sports field  possibly a tennis court with a grassy surface. There are trees in the background  and | The setting is an open grassy field with trees in the background  suggesting a park or recreational area. |
| **Can you describe the main object in the image?** | Yes  the main object in the image is a dog. | Certainly! The main object in the image is a cat. It appears to be sitting on top of a reflective surface. |
| **Can you describe the position of the surfer in relation to the wave?** | The surfer is in the middle of the wave  riding it. The wave is curved and the surfer is positioned in the center of | The surfer is positioned on top of the wave  actively riding it. The wave is curling behind him  and he appears to be surfing |
| **What colors are prominent on the airplane?** | The prominent colors on the airplane are white and blue. | The airplane is predominantly white with blue accents. It also has a blue tail fin that features a red and blue logo. |
| **Given the image  can we tell if it's morning or evening?** | Based on the lighting and shadows in the image  it appears to be late afternoon or early evening. | The image does not provide clear indicators like sun position to determine if it's morning or evening. Therefore  I cannot confidently answer that question based on |

| **Are there any other animals in the picture apart from the bear?** | No there are no other animals visible in the picture apart from the bear. | No there are no other animals visible in the picture apart from the bear. |
|---|---|---|
| **Are there any electronic devices besides the laptop?** | Yes there is a mobile phone visible in the image. | Yes there is a telephone present in the image. |
| **What do you see in the picture?** | The image depicts a scenic landscape with rolling hills and a body of water in the distance. There's a train on the tracks and in the | In the picture I observe a scenic landscape consisting of a river a bridge a bullet train moving on the bridge several buildings in the |
| **What's the relationship between the candle and the igniter in terms of positioning?** | The candle is placed on the left side of the igniter. | The igniter is positioned to the right of the white candle on the coffee table. |
| **If you were to give a brief summary of this image how would you describe it?** | The image depicts a man in a suit and tie standing in front of a green background. He has a serious expression and is holding a cell phone | The image captures a moment of a man in a formal attire with spectacles addressing an audience or participating in a discussion at a formal event |
| **What animals can you see in the image?** | In the image I can see a herd of cows. | I see several bulls in the image. |
| **Describe the weather or atmosphere based on the image.** | The image suggests a clear and sunny day. There are no clouds visible in the sky and the lighting appears to be bright and warm. | Based on the image the weather appears to be cloudy or overcast. The field looks slightly damp possibly suggesting that it has rained earlier |
| **Where is the toilet located in the image?** | The toilet is located in the center of the image. It's positioned on the right side of the image closer to the right edge. | The toilet is centrally positioned in the image situated towards the lower-middle section. |
| **What is the title or theme of the book the older child is holding?** | The title of the book the older child is holding is not clearly visible in the image. | The book the older child is holding is titled "The Lady with the Alligator Purse " and it appears to have illustrations related to alligators |
| **How many people are there in the picture?** | There are three people in the picture. | There are five people in the picture. |
| **Describe the surroundings of the clock tower. What's below or around it?** | Below the clock tower there's a large open area with grass. There are also some trees and buildings in the background. The clock tower is | The clock tower appears to stand atop a larger building which has similar architectural features. Directly below the clock are windows and ledges. The |
| **Based on what you can see what might these riders be doing or where might they be heading?** | Based on the visual content of the image it appears that these riders might be on a leisurely walk or hike in a forested area. | While I cannot deduce their exact intentions from the image alone the riders seem to be journeying through the countryside perhaps for leisure travel |

| **If I wanted to see giraffes in their natural habitat where should I go?** | If you want to see giraffes in their natural habitat  you should consider visiting the savannas or grasslands of Africa. Some popular | To see giraffes in their natural habitat you should consider visiting countries in Africa  particularly in regions such as the East African savannas |
|---|---|---|
| **How many teddy bears are there?** | There are three teddy bears in the image. | There are two teddy bears in the image. |
| **How do you think these apples might be used based on their positioning and surrounding objects?** | Given the positioning of the apples and the presence of the trash can  it's likely that these apples are being discarded or disposed of. They might | Given their positioning on the tablecloth and the presence of kitchen items nearby the apples might be used for eating directly  or they could be ingredients |
| **Can you describe the surroundings of the airplane in more detail?** | Certainly. The airplane is positioned on a snowy ground  and in the background  there's a snowy hill or mountain. There are also some trees | Certainly. The airplane is situated on a sandy beach with the ocean waves visible in the background. The beach appears to be expansive  with the horizon |
| **Which bird is closer to the left side of the image?** | The bird on the right side of the image is closer to the left side of the image. | The bird that is closer to the left side of the image is the one with its beak open. |
| **Describe the bed and its components in the image.** | The bed is a double bed with a white duvet cover. It has multiple pillows some of which are covered with a light-colored sheet. | The bed appears to be a standard hotel bed with a prominent headboard. It's covered with a bedspread and multiple blankets. The bedspread |
| **What's inside the glass?** | Inside the glass  there's a slice of orange. | The glass contains a liquid that looks like a smoothie or a shake with specks possibly seeds or bits of fruit. |

**Table S8. Samples of CorText decoding results on the question-answering task (subject 5, test set).**

| Question | CorText answer | Human caption |
|---|---|---|
| **What are they wearing?** | One of the individuals is wearing a white shirt while the other is wearing a black shirt. | Both chefs are wearing white chef uniforms with hats. One of them has an apron with visible logos and text. |
| **Where is the sun located in the image?** | The sun appears to be located behind the surfer casting a shadow on the water. | The exact location of the sun isn't visible in the image but from the lighting it seems to be shining from above and possibly slightly from |
| **Based on the image is there a picnic happening on the right side?** | Yes there appears to be a picnic happening on the right side of the image. There are picnic baskets and people sitting on the grass. | From the visual content of the image provided there isn't any evident sign of a picnic happening on the right side. |
| **How many gravestones are visible in the image?** | There are multiple gravestones visible in the image. | There are multiple gravestones in the image approximately more than 30. |
| **What about the green onions? How many chopped pieces can you see?** | I can see multiple chopped pieces of green onions scattered across the pizza. | I can see numerous chopped pieces of green onions scattered throughout the plate. There are more than fifteen distinct pieces visible. |
| **What's the position of the business card relative to the coffee cup?** | The business card is located to the right of the coffee cup. | The business card which reads "DROP COFFEE" is placed to the right of the coffee cup. |
| **Do you think the person in the middle is likely to catch the frisbee based on their positioning and hand orientation?** | Based on the image it seems challenging for the person in the middle to catch the frisbee with their current hand orientation and positioning. Their | It's hard to predict the exact outcome but based on the current positioning the person in the middle seems to be in a good position to |
| **Where is the cat positioned in relation to the background?** | The cat is positioned in front of the background which appears to be a cityscape or urban area. | The cat is positioned on top of a car which seems to be parked outside near a residential area with buildings and trees in the background. |
| **How many people are there in the photo?** | There are two people in the photo. | There is one person in the photo. |
| **Can you describe the weather conditions based on the image?** | Based on the image it appears to be a clear and sunny day. | The weather appears to be overcast with a cloud-filled sky. However it doesn't seem to be stormy or rainy. |
| **Is there a clear sky in the background?** | Yes the sky in the background appears to be clear. | The sky is not clearly visible in the image as the primary focus is on the surfer and the wave. However there appears to be no |

| **Can you talk about the natural habitat of such bears and why they might be in the water?** | Certainly! Brown bears like the one in the image are native to North America and Eurasia. They inhabit various habitats including forests | Certainly. The bear in the image looks like a brown bear. Brown bears are often found in forests mountain regions and tundras across |
|---|---|---|
| **Considering the layout of the room do you think it's a spacious room?** | Based on the image the room appears to be moderately sized. There's enough space for the person to sit on the couch and there's | The visible portion of the room suggests it is not very spacious. The bed occupies a significant portion of the visible area and there are items and |
| **Where is the train located?** | The train is located on the right side of the image moving towards the viewer. | The bullet train is located on the bridge that spans across the river. It's roughly at the center of the bridge making its way towards the |
| **On which side of the image is the piano located?** | The piano is located on the left side of the image. | The piano is located on the right side of the image. |
| **Are there any signs of movement or action in the picture?** | No the image appears to be a still shot. There is no indication of movement or action in the picture. | There are no direct signs of rapid movement like motion blur. However the man's posture and facial expression suggest he is actively engaging in a |
| **Besides the bulls are there any other objects or features in the foreground?** | Yes besides the bulls there are some trees and grass in the foreground. The trees provide shade and shelter for the bulls while the grass | Yes there are several logs scattered around the field where the bulls are. |
| **Can you see any audience or spectators in the image?** | No I do not see any audience or spectators in the image. The focus is primarily on the tennis player. | No there aren't any visible audience or spectators in the image. The focus is primarily on the goalkeeper and the immediate surroundings. |
| **Where is the toilet located in the image?** | The toilet is located on the right side of the image. | The toilet is centrally positioned in the image situated towards the lower-middle section. |
| **Is there any interaction between the two children?** | Yes the two children appear to be interacting with each other. The child on the left seems to be pointing at the child on the right | Yes there seems to be an interaction between the two children. The younger child is closely looking at the book while the older child is holding it |
| **How many people are there in the picture?** | There are four people in the picture. | There are five people in the picture. |
| **If there were birds flying around the tower would they appear smaller or larger than the clock's hands?** | If there were birds flying around the tower they would appear much smaller compared to the clock's hands. The clock's hands are quite large | If there were birds flying around the tower they would likely appear smaller than the clock's hands given the scale and size of the clock. |

| **What are the activities these people are involved in?** | The people appear to be engaged in various activities. Some are walking while others seem to be riding the horses. Additionally  there's a person | Both people are riding horses. |
|---|---|---|
| **What kind of diet do giraffes have?** | Giraffes are herbivores and primarily feed on leaves  shoots  and fruits from trees and shrubs. They use their long tongues | Giraffes are herbivores  primarily feeding on the leaves  fruits  and flowers of trees and shrubs. They have a particular |
| **Where is the piano located relative to the teddy bears?** | The piano is located to the left of the two teddy bears. The teddy bear on the left is closer to the piano  while the | The piano is located behind the teddy bears. |
| **Are there any other fruits or vegetables in the image?** | No  there are no other fruits or vegetables visible in the image. | No  the only fruits visible in the image are the apples. |
| **What is the color scheme of the airplane?** | The airplane has a predominantly blue and white color scheme. | The airplane has a white base color with blue and green graphics on it. There's also some text and logos. |
| **Based on their features  can you infer any specific type or species of these birds?** | Based on their features  the birds appear to be some species of antelopes or similar grazing animals. However  without more specific details  it | The birds appear to be Bee-eaters identifiable by their colorful plumage and sleek body structure. Bee-eaters are known for their vibrant colors |
| **What's the design or pattern on the girl's shirt?** | The girl is wearing a shirt with a design or pattern that looks like a cartoon character. The design is colorful and has a playful feel to it | The girl's yellow shirt has some sort of design or emblem on it  but I cannot provide a detailed description of the exact design without more context |
| **Can you count the number of bananas present?** | Yes  there are two bananas in the image. | Yes  there is one banana present in the image. |

**Table S9. Samples of CorText decoding results on the question-answering task (subject 7, test set).**

| Question | CorText answer | Human caption |
|---|---|---|
| **What is in this picture?** | A group of giraffes standing in a field. | TWO SMALL ZEBRAS WALKING AROUND INSIDE OF A GATE. |
| **Give a concise and descriptive caption of this image:** | A close up of a couple of giraffes in a field. | A zebra standing next to a lot of trees. |
| **Describe the important features of the scene in this image:** | Two sheep standing in a field next to each other. | A herd of zebra standing on a lush green field. |
| **Describe the following image in detail in one sentence:** | A pair of brown and white sheep grazing in a field. | A couple of zebra walking across a grass covered field. |
| **Describe the important features of the scene in this image:** | A young boy standing next to a large white horse. | Two zebras walking side by side in an enclosure |
| **What is in this picture?** | A horse is running in a field with trees in the background. | A single zebra grazing in a grass field. |
| **Provide a detailed description of the given image in one sentence:** | A black bear with its tongue hanging out of its mouth. | A zebra standing up by some thin branches. |
| **What is in this picture?** | A close up of a giraffe's face and neck. | A zebra eating grass in front of some small bushes. |
| **Describe the following image in detail in one sentence:** | A large white horse standing in a grassy field. | A zebra is eating from the tall brown and green grass. |
| **What is in this picture?** | A black and white photo of a giraffe in a field. | Zebras grazing on grass in large open area near trees. |
| **Describe the important features of the scene in this image:** | A brown and white donkey standing in front of a stone wall. | A zebra with black and white stripes standing around. |
| **Provide a detailed description of the given image in one sentence:** | A herd of cows grazing in a field. | some zebras standing next to each other and eating some grass |
| **Describe the following image in detail in one sentence:** | Two black and white horses grazing in a field. | A couple of zebra standing next to each other on a field. |
| **What is in this picture?** | A herd of elephants standing in a field. | A large group of zebra on a grass field. |
| **What is in this picture?** | A group of giraffes standing on a dirt road. | Several zebras are huddled together and a log is visible in the background. |
| **Describe the important features of the scene in this image:** | A group of giraffes standing in a field. | A zebra standing around in the middle of dessert. |

| **Give a concise and descriptive caption of this image:** | A giraffe standing in the middle of a field. | A zebra is standing in the grass facing forward. |
|---|---|---|
| **Provide a detailed description of the given image in one sentence:** | Two giraffes standing next to each other in a field. | Two zebras standing head to head in an exhibit. |
| **Describe the important features of the scene in this image:** | Two giraffes standing in a field with trees in the background. | Two zebras in a stone walled enclosure with small trees in it. |
| **What is in this picture?** | A black and white photo of a black and white cow. | Three zebras eating hay from bales inside a barn-like structure. |

**Table S10. Zero-shot caption decoding samples for CorText trained without the zebra cluster (subject 1, test set).**

| Question | CorText answer | Human caption |
|---|---|---|
| **Provide a detailed description of the given image in one sentence:** | A group of people standing on a snow covered beach. | a windsurfer some water a hill sand and some kayaks |
| **Describe the following image in detail in one sentence:** | A man skiing down a mountain with a black and white ski suit. | A man in swim shorts is riding a surf board. |
| **Provide a detailed description of the given image in one sentence:** | A young girl riding a pink bicycle on a dirt path. | A young child riding a wave on top of a board. |
| **What is in this picture?** | A person riding a snowboard down a hill. | A man on a surfboard surfing in the ocean. |
| **Describe the following image in detail in one sentence:** | A person riding a horse on a beach. | This photograph of a surfer entering the ocean is breathtaking. |
| **Provide a detailed description of the given image in one sentence:** | A person jumping off of a snowboard in the air. | A surfer is in mid air while catching a wave. |
| **Describe the following image in detail in one sentence:** | A man skateboarding in a skate park. | A man holding a surfboard stands on rocks and looks down at the ocean. |
| **Describe the important features of the scene in this image:** | A man jumping in the water with a surfboard. | A man on a surf board riding a wave in the ocean. |
| **Describe the important features of the scene in this image:** | A man riding a skateboard down a dirt path. | A man holding a surfboard standing on a wave covered ocean. |
| **What is in this picture?** | A person riding a snowboard in the snow. | A man coming out of the ocean onto the beach holding a surfboard. |
| **Describe the following image in detail in one sentence:** | A black and white photo of a black dog in the snow. | Surfer riding atop a Avery large wave about to jump. |
| **Describe the important features of the scene in this image:** | A man riding a snowboard down a mountain. | a person riding a surf board on a body of water |
| **What is in this picture?** | A deer is walking through the woods near a river. | A skier is skiing alone down the ski run. |
| **What is in this picture?** | A person is riding a snowboard down a snowy hill. | A man on a surfboard on an ocean wave. |
| **Give a concise and descriptive caption of this image:** | A man and a woman skiing down a snowy hill. | a couple of men are laying on surfboards in the water |
| **Give a concise and descriptive caption of this image:** | A person riding a snowboard down a snowy hill. | Surfer getting airborne while riding large wave in ocean current. |

| **Describe the following image in detail in one sentence:** | A person snowboarding down a snow covered hill. | A young surfer surfboarding in the ocean doing tricks |
| --- | --- | --- |
| **Describe the following image in detail in one sentence:** | A group of people skiing down a snowy hill. | A group of people surfing with wave in the background. |
| **Give a concise and descriptive caption of this image:** | A man standing in a field with a golf club. | a man riding a wave on a surfboard. |
| **What is in this picture?** | A boat docked in a marina with a house in the background. | White car parked in a lot with two surfboards on top. |

**Table S11. Zero-shot caption decoding samples for CorText trained without the surf cluster (subject 1, test set).**

| Question | CorText answer | Human caption |
|---|---|---|
| **Describe the following image in detail in one sentence:** | A person is riding a white boat in the water. | A small propeller plane is at a sandy beach. |
| **Describe the following image in detail in one sentence:** | A car parked in front of a building with a large sign. | The green and red airplane is sitting on the runway. |
| **Describe the important features of the scene in this image:** | A motorcycle is parked next to a car. | a blue korean air airplane sitting on the track |
| **Provide a detailed description of the given image in one sentence:** | A blue and white striped tennis court with a tennis net. | An Air Pacific jump jet in flight in the air. |
| **What is in this picture?** | A motorcycle is driving down a road with a city in the background. | a blue and yellow plane is in the sky |
| **What is in this picture?** | A bird flying in front of a large windmill. | A large propeller airplane flying through a cloudy sky. |
| **Describe the following image in detail in one sentence:** | A person riding a skateboard down a hill. | A passenger plane is taking off from a runway |
| **Describe the following image in detail in one sentence:** | A blue and white train is sitting on the tracks. | A white jet with blue and orange detailing sits on the runway. |
| **What is in this picture?** | A group of people flying kites in the sky. | A group of planes flying in the sky letting out smoke |
| **Provide a detailed description of the given image in one sentence:** | A white dog sitting on a bench next to a car. | a large jet airplane parked on a runway with people surrounding it |
| **Describe the important features of the scene in this image:** | A white boat is docked at a marina. | a couple of large airplanes that are parked in a parking lot |
| **Describe the following image in detail in one sentence:** | A train is sitting on the tracks near a building. | A large Thai jetliner parked on an airport tarmac. |
| **What is in this picture?** | A close up of a motorcycle parked on the side of the road. | A passenger jet airliner with the word "AIRFRANCE" is in the air in front of a cloudy sky. |
| **Describe the following image in detail in one sentence:** | A group of police officers standing in front of a police car. | a couple of guys that are walking to a plane |

| **Provide a detailed description of the given image in one sentence:** | A train is sitting on a track in a field. | A plane on the runway under cloudy skies. |
|---|---|---|
| **Describe the important features of the scene in this image:** | A motorcycle is parked on the side of the road. | Airplanes are parked on the tarmac of an airport. |
| **Describe the important features of the scene in this image:** | A large white boat on the water with a large crowd of people watching it. | An airplane sits on a runway in the middle of snowy grass. |
| **Provide a detailed description of the given image in one sentence:** | A row of parked cars on a street. | A jet airplane flying above some city buildings. |
| **What is in this picture?** | A group of people sitting in the back of a truck. | Several buses are parked on a tarmac near some planes. |
| **Describe the important features of the scene in this image:** | A motorcycle parked on the side of the road. | A large passenger jet flying through the blue sky. |

**Table S12. Zero-shot caption decoding samples for CorText trained without the airplane cluster (subject 1, test set).**

| CorText w/o Zebra answer |
| --- |
| I see a zebra. |
| The noun from the category "zebra" is present in the scene. |
| I see zebras in this scene. |
| The scene features a zebra. |
| The scene features a zebra. |
| The scene features a zebra. |
| The noun from the category "zebra" is present in the scene. |
| The scene contains a zebra. |
| The scene features a zebra. |
| The scene contains a zebra. |
| The scene features a zebra. |
| The scene features a group of zebras. |
| The scene features a zebra. |
| The scene features a group of zebras. |
| The scene features a group of zebras. |
| The scene features zebras. |
| The scene contains a zebra. |
| The scene features a zebra. |
| The scene features a zebra. |
| The scene features a zebra. |

**Table S13. Zero-shot choice experiment for CorText trained without the zebra cluster (subject 1, test set).**

| CorText w/o Surfer answer |
| --- |
| The category of "surfer" is relevant to this scene. |
| The noun from the category "surfer" is present in the scene. |
| I choose "surfer". |
| There is no zebra  airplane  or surfer in the image. |
| The category of the noun in the scene is: surfer. |
| surfer |
| I don't see a zebra  airplane  or surfer in the image. |
| The chosen noun is: surfer |
| The category of the noun in this scene is: surfer. |
| The chosen noun from the categories is "surfer". |
| I choose "surfer". |
| The following noun: surfer |
| The category of the noun in this scene is: zebra. |
| The chosen noun is: surfer |
| The category of "surfer" is relevant to the scene. |
| The one noun from the following categories that is in this scene is: surfer. |
| The category of "surfer" is relevant to this scene. |
| The chosen noun is: surfer |
| There is no zebra  airplane  or surfer in the image. |
| The category of "airplane" is relevant to this scene. |

**Table S14. Zero-shot choice experiment for CorText trained without the surf cluster (subject 1, test set).**

| CorText w/o Airplane answer |
| --- |
| Airplane |
| Airplane |
| The airplane. |
| The scene contains a zebra. |
| The airplane. |
| Airplane. |
| Surfer |
| Surfer |
| Surfer |
| Surfer |
| Airplane |
| The airplane is in the sky. |
| There is a zebra. |
| I choose "zebra". |
| The scene contains a zebra. |
| The airplane is in the scene. |
| There is a zebra. |
| The airplane is in the scene. |
| There is a zebra. |
| Surfer |

**Table S15. Zero-shot choice experiment for CorText trained without the airplane cluster (subject 1, test set).**

| $\beta = 0$ (no stimulation | $\beta = 0.15$ | $\beta = 0.25$ | $\beta = 0.5$ | $\beta = 1$ | $\beta = 5$ |
|---|---|---|---|---|---|
| A bowl of green salad with various vegetables and fruits. | A bowl of food with vegetables and a fork. | A plate of food with a piece of broccoli and some green beans. | A woman holding a pizza smiling at the camera. | A man and a woman both wear glasses. | Two men standing in a car smiling and holding a camera. |
| A large airplane flying in the sky with clouds underneath it. | A yellow and white airplane with a man on the ground. | A man is on a boat while looking at a train. | A man with a mustache is sitting in a boat. | An older man and a young boy smiling together on a dock. | A man and woman with a dog in a boat. |
| Two large black and white bears in the wild. | A man wearing a hat is holding a dog in a lake. | A man with a beard and a hat | a man and woman smiling at the camera | a man and woman on a dock next to a boat | A couple smiling in the snow near a snow covered car. |
| A room with a bed and a desk with a computer and chair. | A group of people sitting around a table with a fire pit in the background. | A family of people sitting in a room filled with toys. | A couple sitting next to a woman in a car. | A man and a woman are on a boat. | Two men standing in front of a car with a baby. |
| a person riding a motorcycle in the sand | Two people on a motorcycle in the countryside. | A couple of people standing on top of a hill. | A couple standing in front of a red car. | A man and a woman posing in a car. | A couple of people standing in front of a car. |
| A small airplane flying over a large boat in the water. | A boat in the ocean with a dog on it. | A dog on a boat in the ocean. | A man and woman in a boat on the ocean. | A man and a woman on a boat with a dog. | Two men wearing sunglasses stand by a boat. |
| A living room with a blue couch and a flat screen TV | A woman sitting on a couch with a dog. | A couple sitting on a couch with a dog. | A man and a woman are sitting together in a car. | Two people sitting next to each other at a table. | A couple smiling in front of a camera. |
| A red and white bus in a parking lot. | A bus with a sign on the front that reads "First" | A woman and man are standing next to a red bus. | A woman and a man standing next to a train. | A couple smiling and holding a camera in a car. | A man and woman smile at the camera. |
| A boat is sitting on a dock on a beach. | People on a boat near a dock with a boat ramp. | A boat in a harbor with a man and a woman on it. | Two men sitting on a boat in the water. | Two men sitting on a boat on a body of water. | A man and a woman posing together at a table. |

| A green truck and a green car driving down a dirt road. | A black and white photo of a man in a black and white truck | A woman with a hat is standing next to a train. | An older couple that are standing outside next to a car. | A man and woman are on a boat in the water. | Two men standing together smiling and looking at the camera. |
|---|---|---|---|---|---|
| A glass vase on a wooden table with a mirror. | A photo of a clock on a wall near a plant. | a photo of a clock in a flower pot | a couple of people in a picture together | A couple of people standing in front of a dog. | A woman and a man sitting together in a car. |
| A sign that reads "No Parking" and a green traffic light. | a man standing next to a stop sign at a bus stop | A man sitting in a convertible on top of a hill. | A man and a woman are on a boat in the water. | Two men sitting on a bench outside of a building. | a couple of people are smiling at the camera |
| A beautiful kitchen with many different items. | A couple of people sitting at a table with a fire in the fireplace. | A couple sitting on a couch and smiling. | A man and a woman are sitting next to each other. | A smiling man and woman with a dog. | Two men sitting next to each other in a boat. |
| A plate of pizza with some green leaves. | A person holds a pizza with a green sauce. | A pizza with green sauce and red sauce on top of the pizza. | A couple of people sitting next to a pizza. | Two people smiling in front of a camera. | Two people in a car smiling at each other. |
| A zebra standing in the middle of a field. | a white and brownish colored sheep that is walking in the grass | A man and his dog walking on a dirt road. | A man and a woman in a field with a dog. | Two men stand outside a building wearing a hat and a scarf. | a man and a woman sitting on a bench next to each other |
| a close up of a giraffe in a zoo | A large giraffe with a black and white pattern. | A close up of a black and white cat looking up. | A man wearing a hat and a woman standing behind him. | An elderly man wearing a tie and a hat. | An older man and a young girl wearing glasses. |
| A sail boat flying high above the ground. | A couple of people are flying a kite on a beach. | A man and a woman are flying a kite on a windy day. | A man and woman standing near a boat in the water. | Two people on a boat on a lake. | Two men with glasses sitting at a table outside. |
| A wooden bench and a table with a flower pot and a book on it. | a man sitting at a picnic table next to a boat | a couple sitting on a boat next to a dog | A man and a woman sitting in a boat. | Two people sitting in a boat with one of them wearing a hat. | An older couple sitting on a bench smiling. |
| A cat that is standing in a vase with a red and white | A dog with a bow tie on its neck. | A man with a tie and a hat is standing near a cat. | A man and woman holding a cat in a pink collar. | A man and woman sitting in a boat with a dog. | A man and woman wearing a bow tie and glasses. |
| checkered background. | | | | | |
| A bunch of bananas, tomatoes, and broccoli sitting on a table. | A black and white dog with a purple collar on a table next to a book. | A man and a woman sitting next to each other with a dog. | a man and woman holding a cat on a table | Two men with a dog sitting in front of a window. | A couple smiling for a picture outside. |

**Table S17. Microstimulation experiment: inducing people predictions.** Predicted captions for microstimulation experiment by CorText trained on subject 1 with increasing $\beta$ reveals that microstimulation induces people to be mentioned in the caption for trials in the test set for which there are no people present.

| $\beta = 0$ (no stimulation | $\beta = -0.15$ | $\beta = -0.25$ | $\beta = -0.5$ | $\beta = -1$ | $\beta = -5$ |
|---|---|---|---|---|---|
| A man riding a surf board on top of the water. | A person riding a white surfboard in a river. | A man riding a horse through the water. | A man riding a horse through a field of green grass. | A person is standing on top of a blanket that is on the ground. | A room with a bed and a chair. |
| a man and woman stand on the beach with their surf boards | A person is standing on a surfboard on the beach. | A person on a surfboard on the beach. | A person sitting on a surfboard on a beach. | A person is sitting on a surfboard on the sand. | A bed with a white comforter and a few pillows on it. |
| A group of kids playing in a field with a frisbee. | A group of children playing on the beach with a frisbee. | A group of people are playing in a field with a frisbee. | A person in a red shirt is playing with a frisbee. | A man is sitting on the grass with a frisbee in hand. | A living room with a white couch, a coffee table, a bed and a vase. |
| A man is sitting on a bed and holding a remote control. | A man is sitting on a bed playing a game on his computer. | A woman laying on a bed with her laptop open. | A living room with a TV on a couch | A bedroom with a bed, a lamp, and a chair. | A bed room with a bed, a chair and a table. |
| A person on a bike is in a park. | A man playing with a frisbee on the sidewalk. | A man is running with a skateboard while holding a baseball bat. | A man rides a motorcycle on the sidewalk. | A person stands in a room holding a blue motorcycle. | A room with a white table with a vase and two chairs. |
| A young boy is laying on top of a large blanket. | A girl laying on a bed with a bike on top of her. | A child sitting on a bed with a bunch of skateboards. | A person standing on a bed with a surfboard. | A collection of clothes on a bed with a backpack. | A room with a bed and a few chairs |
| A group of people standing outside with flags. | A group of people on horses in the rain. | A group of people are standing around a bike with a jersey on. | A group of motorcycles are parked in front of a building. | A few motorcycles parked next to a banner. | a room with a bed and some furniture |
| a couple walking through the woods on a sunny day | A group of three men walking down a dirt road. | A group of sheep grazing in a field. | a horse is walking in a room with some furniture | a person is sitting on a bed with a guitar in the background | A room with a bed and a table with a vase in the middle. |
| A little girl sits on a bed next to a man in a bed. | A young boy is laying on a bed in a room. | A child laying on a bed with a red blanket. | A bed with a quilt and pillows is in a room. | A bed with a quilt and a small lamp. | A room with a bed and a dresser with a mirror on it. |

| A young man with a baseball bat in his hands. | A young man is playing baseball in the outfield. | A man in a baseball jersey swings a baseball bat. | A person playing a game of tennis in a park. | A person is sitting on a bed in a room. | A large bed with a red blanket and pillows. |
|---|---|---|---|---|---|
| A man is holding a frisbee while walking in a park. | A man playing with a frisbee in a field. | A person holding a baseball bat while playing a game. | A person playing with a frisbee on a field. | a person is playing with a baseball bat | A room with a bed and a few chairs. |
| A group of people playing soccer on a field. | A group of kids playing soccer in the park. | a group of children playing soccer in the park | A person riding a bicycle on a grass field. | A person riding a bike through a park. | A bed with a few pillows and a blanket is in a room. |
| a person is flying a kite on the beach | A person is standing in a field with a kite. | A person walking with a kite in a park. | a person standing next to a bicycle in a field | A person riding a bike with a large banner on the back. | A bed with a few pillows and a couple of chairs in a room. |
| A woman stands near a toilet in a room. | A person standing in front of a white cabinet with a red chair next to it. | A person walking through a room with a bed and a chair. | A bed in a room with a fan on the wall. | A room with a bed, a dresser, and a window. | A bed with a white quilt and a black and white vase. |
| A man walking down the street with a bicycle. | A person on a motorcycle riding on the side of the street. | A black and white photo of a horse riding in a street. | a white and red bike parked on a cobble street | A room with a bed and a couple of chairs. | A room with a bed, a window and a vase of flowers. |
| A person sitting on a bench with a laptop. | A person is sitting on a bench with a skateboard next to them. | a couple of people sitting at tables with laptops | A bed and chair in a room with a few things on the bed. | A bed with a blanket and a lamp on the table. | a room with a bed and a table with some things on it |
| A person on a surfboard riding a wave in the ocean. | A person is riding a surfboard on top of a wave. | A man is surfing in the ocean with a red board. | A person is standing on a surfboard in the ocean. | A person sitting on a bed with a surf board next to them. | A bed with a blue blanket and a few pillows. |
| A woman swinging a tennis racket in a park. | A man holding a tennis racket in his hand while standing on a court. | A man in a blue shirt holding a tennis racket on a green court. | a man standing on a tennis court holding a racket | A person in a white jersey playing with a frisbee. | a large bed with a green blanket and white curtains |
| A man playing with a frisbee in a field. | A man is hitting a frisbee with a bat. | A man playing a baseball game in a park. | A man is standing with a baseball bat. | A person is standing in a room with a bike. | A bed room with a bed, a dresser and a window. |

| A group of people sitting at a table with a white table cloth. | A man sits at a table with two people next to him. | a person sitting in a room with a guitar and a chair | A man sitting on a bench with a guitar. | A bedroom with a bed and a dresser. | a room with a bed, , a chair, and a table |
|---|---|---|---|---|---|

**Table S18. Microstimulation experiment: reducing people predictions.** Predicted captions by CorText trained on subject 1 for microstimulation experiment with decreasing $\beta$ reveals that for trials in the test set with people present in the images, decreasing inhibition causes the model to predict decreasingly fewer people in the caption.

| $\beta = 0$ (no stimulation | $\beta = -0.15$ | $\beta = -0.25$ | $\beta = -0.5$ | $\beta = -1$ | $\beta = -5$ |
|---|---|---|---|---|---|
| A person is cutting a cake while another person looks on. | A man is cooking a pizza in the kitchen with his wife. | A man is using a whisk to mix batter in a bowl in a kitchen. | A man is working on a food truck in a garage. | A man is using a blow torch to melt cheese on a piece of metal | A man is riding a motorcycle that has a red stripe on the side. |
| The image showcases a wooden cutting board with sliced vegetables. The vegetables | A plate of stir fried vegetables with rice. | The image showcases a plate of food consisting of vegetables and a sauce | A tray of food that has various kinds of meat. | The image displays a group of motorcycles parked on a street. Some | A green and black motorcycle parked next to a blue bus. |
| Several cows graze in a grassy field with some trees. | A large group of zebras are standing on a large grassy plain. | There are several animals in the image. Some are standing and some are | A group of zebras grazing in a field. | The image showcases a herd of zebras in a grassy field. | A white and blue truck is parked on a road. |
| The picture shows a living room with two sofas a coffee table and several items on | I see a room with a white wall as the primary background. There are several objects on the table and floor including a laptop computer | A room with a desk sofa and various other furniture pieces. | A room with furniture and a computer and a bookshelf. | a bunch of different types of bikes in a room | There are several motorcycles in this picture. |
| A group of motorcycles are driving on a dirt road near a body of water. | a group of motorcycles are driving across a dry dirt road | a herd of cows on a dirt road | A motorcycle is on a dirt road. | A white and red motorcycle is on the road. | A black and white motorcycle is on a road. |
| A young boy is playing catch with his sister in the park. | Two women on the beach playing a game of frisbee. | A group of kids are playing soccer on a field. | A group of young women on a beach with one of them on her | A young girl and a boy on a field with a soccer ball | A man is holding onto the back of a red motorcycle while riding it. |
| The image showcases a cozy living room setting. It has a large | The image depicts a cozy bedroom. There is a single bed in the | The picture shows a room with a large wooden desk in the center. On the desk there | The image features a cozy living room with a wooden furniture set. The | A blue and white train sitting on a rail track in a room. | A red and white train parked in a garage. |
| A white bathroom with a toilet and shower. | A toilet toilet paper roll and a trash can in a bathroom. | A white toilet and sink are in a small bathroom. | A toilet with a broken seat and a trash can next to it. | A blue and white toilet is sitting on a cement floor in a room. | A black and white train sitting on the tracks. |

| The image features a clock tower with a clock face prominently displayed. The | A large clock tower with a clock face on it. | A large white motorcycle parked next to a red brick building with a clock on it. | A black and white bicycle with a basket and a sign on it. | A vintage motorcycle parked next to a wooden bench. | There is a red and white train parked on the tracks. |
|---|---|---|---|---|---|
| A banana with a knife in the middle. | A banana with a bite taken out of it and a fork on the | a banana and a spoon sitting on a table | A banana in a bag on a table. | a close up of a hot dog on a plate with a fork | A motorcycle is parked on the sidewalk outside a building. |
| A vase of flowers sits in a window with a mirror reflection. | A glass vase with flowers in front of a mirror. | The image features a vase placed on a table with a glass vase | A glass vase with a picture of a flower on a table. | The image displays a close-up of a motorcycle. | A small blue and white train on a white background. |
| A street scene with a sign for a city's parking meter. | A large stop sign that has a white and red stripe on it. | A street sign with a street name that reads 'Hawthorne Ave' and a traffic light are in the image. | A train sitting on the tracks with a city in the background. | a white bus on a road in front of a building | A black train that is parked on the side of the road. |
| A room with a table and some lamps. | A room with a table and chairs. | a kitchen with a table and chairs in it. | a black car and a blue car with a window with a view of a | A couple of old cars sitting on a driveway. | A truck is sitting on a parking lot with a car behind it. |
| A pizza that is cut into slices is on a plate on a | A pizza is sitting on a table with some green stuff on it. | A pizza that has been cut up into pieces. | A plate of pizza and a drink on a wooden table. | The image depicts a white car with a wooden body parked on a | A white and yellow truck in a parking lot. |
| A zebra stands in a field with some trees and hills in the background. | a zebra is standing in the middle of a field of green grass | A zebra standing in the grass with the sun setting in the background. | a zebra grazing in a field | A zebra standing on a dirt road with grass and trees | a white and red zebra on the dirt road |
| The image showcases a giraffe with a light-colored coat. | A giraffe stands in a field with a tree in the background. | The image shows a giraffe in a natural outdoor setting. The | A close up of a giraffe with its head in the grass | A large black and white striped zebra in a grass field. | The image depicts a yellow and black striped train car. |
| a large kite flying above the ocean. | The image showcases a serene sky with a few clouds. | A plane in flight with a beautiful landscape below it. | A blue airplane flying over a field. | A small airplane is on the runway. | A train is parked on a street in front of a building. |
| The image features a room with a wooden table in the center. | A wooden bench is placed near a table with a red umbrella and a chair. | The image shows a room with a table and chairs. The table has | A bench sitting next to a bike on a deck. | A bicycle that is parked next to a wooden bench. | The image showcases a vintage motorcycle parked on the street. The motorcycle |

| A close up of a stuffed bear sitting on a white bed. | A small bear made of white and black felt sitting in a chair. | A small white teddy bear with a pink bow. | A teddy bear is sitting on a white wooden chair. | A brown and white teddy bear sitting on a white chair. | A red and white pick-up truck is parked on a dirt road |
|---|---|---|---|---|---|
| a table with a bag of carrots and a bottle of salsa on it | The image shows a table with a few objects on it. | There are multiple items in the image. A toy teddy bear is | In the image I see a wooden bench with a few objects on it. | The image showcases a collection of vehicles. There are three vehicles visible: | A black and white motorcycle with a license plate. |

**S19. Microstimulation experiment: inducing vehicle predictions.** Predicted captions for microstimulation experiment by CorText trained on subject 1 with increasing $\beta$ reveals that microstimulation induces vehicles to be mentioned in the caption for trials in the test set for which there are no vehicles present.

| $\beta = 0$ (no stimulation | $\beta = -0.15$ | $\beta = -0.25$ | $\beta = -0.5$ | $\beta = -1$ | $\beta = -5$ |
|---|---|---|---|---|---|
| A commercial airliner is taxiing on the runway at an airport. | A large airplane that is sitting on the runway. | An airplane with a blue and white body with a red stripe. | The image features an airplane with the logo of Lufthansa flying in the sky. The airplane is in mid-air with a blue sky. | A tall white church clock tower next to a large green grass field. | The image features a city street with a tall ornate iron lamp post in the center. |
| The image depicts a body of water with a few boats in it. In the background | I see a large boat docked in a body of water. There | A ferry boat on a body of water with mountains and buildings. | The image features a waterfront area with a pier extending into a body of water. | A view of a city with a large clock tower and boats on | The image showcases a city street scene with a green grassy area on |
| A small airplane is flying over the ocean. | A small airplane flying above the water. | A large airplane flying over water with a boat in the foreground. | The image captures a sailboat flying a large red and white ' | A kite is flying over a person on a beach. | A sign in the middle of a green grassy field with a bench |
| A red double decker bus parked on a street in front of a building. | The image features a bus with a white and red color scheme. | A red bus driving down the street in front of a row of shops. | A bus is parked next to a building with a sign that says 'Coles.' | A small white church in front of a row of houses. | A person is sitting on a bench in front of a flower bed with a fountain in the background |
| a boat is on the beach next to a building | A boat sitting on a beach with a small building in the background. | A boat docked in the water with a beach in the background. | A person is sitting on a beach with umbrellas boats and a beach hut in the background. | The image showcases a beach setting with several beach chairs and umbrellas. | A man standing next to a table with umbrellas and chairs. |
| The picture depicts a rustic rural landscape with a large old-fashioned red and white water tower | A small town with old buildings and a river in the background. | A large red and white building is standing on a hill overlooking a valley. | A large stone wall and a fountain are in a green grassy area. | The image displays a serene outdoor setting with a fountain in the foreground. There are several stone | A woman stands next to a fountain in a park with a few people walking in the background. |
| A large white airplane with a red stripe on the side sits on the tarmac. | A large green and white airplane sits on the tarmac. | A street scene with a big green and white traffic sign on a pole. | A sign that reads 'GO' is at the end of a road. | A green stop sign with a crosswalk. | The image depicts a park setting with a tree in the background. |

| A city street with cars parked along the curb and a sign with an | The image captures a street scene with multiple cars trucks and a streetlight situated on a street with a sidewalk | A street view of a city with traffic lights and a car. | A street scene with a parked car and streetlights. | A city street scene with several street signs a fire hydrant and a street lamp. | In this image I see a park scene. There's a bench where a person sits. |
|---|---|---|---|---|---|
| A parked motorcycle with a large load of flowers. | The image features a parked motorcycle with a helmet and a basket attached to it. The motorcycle | A bicycle is parked near a fire hydrant in an urban setting. | The image features a vibrant city scene. There is a fire hydrant | A park with a bench and a flower pot. | A park with a water fountain in the center. |
| a city street with cars and a person on a bike | A bike is riding on the sidewalk near a building. | A city street with a person riding a skateboard and another person walking. | A man walks down the street while carrying a large backpack. | a man walking on a sidewalk with a kite | A person walking in a park with a large tree in the background. |
| A train travels through a desert like landscape. | A train is moving down a track through a city area with buildings. | A train is sitting on the tracks in a field. | A clock tower stands tall in front of a train station. | The image showcases a street scene with a clock tower in the background. | The image features a green grassy field with a clock tower in the |
| A group of horses running on a snowy mountain. | A beautiful landscape of mountains and water. | A snow covered mountain range with a lake in the foreground. | A view of a lake with mountains in the background. | A green grassy area with a large umbrella on it. | A bench sitting next to a sign on a grassy hill. |
| A person riding a skateboard down a street lined with buildings. | A person on a skateboard in front of a building. | A man is walking on a street with a skateboard and a backpack. | A young man walking down the street with a skateboard. | A person walking down a sidewalk in front of a building. | A man is walking in a park with a large umbrella. |
| The image features a red motorcycle parked on the beach. It is partially | A bike is parked on the beach with a surfboard on top of it. | The image features a beach scene with a bicycle parked on the sand. | A red bicycle is parked on the beach. | The image features a beach scene. There's a person sitting on a surfboard with | A bench with flowers is placed on the sidewalk. |
| A bus is traveling down a street with a tree in the | A city bus is traveling down the street with a city building in the | A double decker bus driving on a city street. | A city bus driving down the street with a tree in the background. | A city street with a stop sign and a red bus. | A woman is standing next to a stop sign in front of a park. |
| A large double decker bus is driving through a city. | A yellow double decker bus driving on a street. | A double decker bus is traveling down a road. | A red bus driving down a road lined with buildings. | A red school bus is parked on the street. | A bench sits in a public square near a water fountain. |

| A person riding a sailboard in the ocean with a dog running next to him. | A group of people standing on a beach with a sail boat in the water | There are two people in the image. One of the individuals is holding a kite and the other person is standing on the beach. There are | A woman in the water with a dog on a beach. | A group of people standing on the beach with some of them flying kites. | A person sitting on the grass in a park with a sign that says 'PARKING PRO’. |
|---|---|---|---|---|---|
| A group of people on horseback are riding in a park. | A group of people riding their bikes on a street. | People riding bikes down a street in front of buildings. | A group of people riding bikes on a dirt path. | People sitting on benches in a park. | A woman sitting on a bench in the park. |
| A red airplane with a propeller flying over a field with trees. | A red airplane is flying in the air. | A plane flying in the sky with a red and white stripe on the | A large white airplane flying over a green grass field. | A person walking with an umbrella in a field with a tree. | The image depicts a garden setting. In the center of the image |
| A bus with a blue and white sign is parked next to a building. | A red bus with a black stripe is parked next to a building | A red double-decker bus is parked in a street in front of | A bus is parked in front of a building with a clock on it. | A clock tower in the distance with a church in the background. | A street with a flower pot and a few people walking. |

**Table S20. Microstimulation experiment: reducing vehicle predictions.** Predicted captions by CorText trained on subject 1 for microstimulation experiment with decreasing $\beta$ reveals that for trials in the test set with vehicles present in the images, decreasing inhibition causes the model to predict decreasingly fewer vehicles in the caption.